\documentclass{article}
\usepackage{graphicx}
\usepackage{lmodern}
\usepackage{booktabs}
\usepackage{enumerate}
\usepackage{bbm}
\usepackage{comment}
\usepackage{hyperref}
\usepackage{amsmath}
\usepackage{amssymb}
\usepackage{amsthm}
\usepackage{authblk}
\usepackage{natbib}
\usepackage{algorithmic}
\usepackage{algorithm}
\usepackage{xcolor}
\usepackage{mathtools}
\usepackage{wrapfig}
\usepackage{nicefrac}
\usepackage[margin=1in]{geometry}
\usepackage[capitalize,noabbrev]{cleveref}
\setcitestyle{authoryear,round,citesep={;},aysep={,},yysep={;}}

\usepackage[utf8]{inputenc}
\usepackage[T1]{fontenc}   
\usepackage{url}           
\usepackage{amsfonts}      
\usepackage{nicefrac}      
\usepackage{subcaption}
\usepackage{ifthen}
\usepackage{siunitx}
\usepackage{enumitem}
\usepackage{tikz}
\usepackage{bbold}

\usepackage{commands}

\title{How Width and Data Shape Generalization Scaling Laws in Quadratic Neural Networks}
\date{}

\author[1]{Julius Girardin\thanks{These authors contributed equally to this work.}}
\author[1]{Emanuele Troiani$^\star$}
\author[1,2]{Yizhou Xu$^\star$}
\author[1]{Vittorio Erba}
\author[2]{\\ Florent Krzakala}
\author[1]{Lenka Zdeborová}

\affil[1]{\small Statistical Physics of Computation Laboratory, \'Ecole Polytechnique F\'ed\'erale de Lausanne (EPFL)}
\affil[2]{\small Information, Learning and Physics Laboratory, \'Ecole Polytechnique F\'ed\'erale de Lausanne (EPFL)}

\newtheorem{result}{Result}

\usetikzlibrary{hobby} 
\usetikzlibrary{arrows.meta} 
\tikzset{>={Latex[length=4,width=4]}} 
\usetikzlibrary{calc,intersections,decorations.markings}

\colorlet{mylightblue}{blue!20}
\colorlet{myblue}{blue!50!black}
\colorlet{mydarkblue}{blue!30!black}
\colorlet{mylightred}{red!10}
\colorlet{myred}{red!50!black}
\colorlet{mydarkred}{red!60!black}
\colorlet{mydarkgreen}{green!30!black}
\tikzset{
  midarr/.style={decoration={markings,mark=at position #1 with {\arrow{stealth}}},postaction={decorate}},
  midarr/.default=0.5
}

\begin{document}
\maketitle
\begin{abstract}
Understanding how performance scales jointly with model size and data is a central problem in modern machine learning. Existing theoretical works on scaling laws typically describe generalization as a function of data or compute, often in fixed-feature or infinite-width regimes and for online SGD. Here, we instead study how generalization scales with the number of trainable parameters and the number of samples in a feature-learning model.
We analyze $\ell_2$-regularized empirical test error minimization in a quadratic two-layer network in a finite-sample setting with structured data. This setting allows for an explicit characterization of the generalization error as a function of the number of samples, model width, and regularization. Our results reveal a phase diagram with distinct scaling regimes as the number of parameters varies. In particular, the generalization error follows data-dependent power laws controlled by the spectral structure of the target. We further characterize the transitions between regimes, including the onset of interpolation, and their impact on generalization. 
\end{abstract}
\section{Introduction}
\label{sec:intro}

Scaling laws have become a central tool for understanding modern machine learning systems, relating performance to the amount of data, model size, and compute \citep{kaplan2020scaling,brown2020language,hoffmann2022empirical}. On the theoretical side, a growing body of work has obtained precise scaling descriptions of generalization as a function of the number of samples or compute, see e.g. \citep{caponnetto2007optimal,bordelon2020spectrum,cui2021generalization,bahri2024explaining,paquette2024four,defilippis2026scaling,defilippis2026optimal,cagnetta2026deriving}. A separate line of research studies how increasing the number of parameters affects optimization, initialization, or expressivity, without directly addressing data-dependent generalization \cite{yang2021tuning,bordelon2023depthwise,chizat2024feature,chaintron2026resnets}. Works that do analyze generalization as a function of model size typically focus on regimes with fixed representations, such as random feature models \cite{spigler2020asymptotic,maloney2022solvable,bahri2024explaining,atanasov2024scaling,bordelon2024dynamical,defilippis2024dimension} where increasing the number of parameters does not alter the learned features, and the role of model size is therefore fundamentally limited. 

Studies of scaling laws in model size in the feature learning regime were initiated only very recently \cite{ren2025emergence,arous2025learning}. 
In these works, the scaling laws are formulated for idealized dynamical limits, where performance is controlled by finite training time, or by the product of learning rate and number of iterations. A characterization of the scaling laws of the test error, train error, and their difference, the generalization gap, as a function of the number of samples and the model size in a feature learning setting remains open.

In this work, we fill this gap by characterizing how the generalization error of the empirical risk minimizer scales with the number of parameters and samples in a feature-learning setting. This allows us to disentangle their roles and characterize how data availability, width, regularization, and target structure jointly determine the performance of the trained predictor. To make this tractable, we consider a minimal yet non-trivial model: a quadratic two-layer neural network trained on high-dimensional data with a power-law structure. This setting captures a genuine feature-learning regime while remaining amenable to theoretical analysis. Our goal is not to model realistic architectures, but to isolate the mechanisms by which model size, data, and data structure interact to determine generalization. \looseness=-1

\paragraph{Model Setting.}
We consider the $\ell_2$-regularized empirical risk minimization (ERM) problem 
\begin{equation}
\label{eq:def:erm}
    \bhW = \underset{\bW}{\min} \,\,  \caL(\bW) 
    \quad\text{where}\quad 
    \caL(\bW) = \sum\limits_{\mu=1}^{n} \left(y_{\mu}-f_p(\bx_{\mu};\bW)\right)^{2} + \lambda ||\bW||_{\rm F}^{2}
    \, ,
\end{equation}
for the class of quadratic two-layer neural networks with $p$ hidden units
\begin{equation}
\label{eq:def:student}
    f(\bx;\bW) 
    = \frac{1}{\sqrt p} \sum_{j=1}^{p}
    \sigma_j\left(\frac{\bw_{j}^{\top}\bx}{\sqrt{d}}\right) 
    = \frac{1}{\sqrt p} \sum_{j=1}^{p}
    \left(\left(\frac{\bw_{j}^{\top}\bx}{\sqrt{d}} \right)^2 - \frac{||\bw_{j}||^{2}_{2}}{d}\right) 
\end{equation}
where $\bW = (\bw_1 | \dots | \bw_p)^\top \in\mathbb{R}^{p\times d}$ are the first-layer trainable weights and we fixed the second layer weights to one. The activation $\sigma_j$ is a centered quadratic function.
In this work, we are interested in characterizing the properties of global minimizers of the objective in \eqref{eq:def:erm}. 
For this purpose, the interplay between the network width $p$ and the input dimension $d$ plays a central role. 
Indeed, for the centered quadratic activation in \eqref{eq:def:student}, the network can be rewritten as a linear model with structured data and weight re-parametrization:
\begin{equation}
\label{eq:def:studentS}
    f(\bx_\mu;\bW) 
    = {\rm Tr}\!\left[\bS \, \bG_\mu\right]\, , 
    \qquad 
    \bS = \bW^\top \bW/\sqrt{pd} \in \bbR^{d\times d}\,,
    \qquad 
    \bG_\mu = (\bx_\mu\bx_\mu^{\top} - \bI_{d})/\sqrt{d}.
\end{equation}
As $\bW$ varies over $\bbR^{p\times d}$, the matrix $\bS$ ranges over the cone of positive semidefinite matrices with rank at most $p$. 
Therefore, minimizing over the network weights is equivalent to solving the constrained optimization problem
\begin{equation}
\label{eq:def:erm_equiv_intro}
    \bhS 
    = \underset{\bS \succeq 0 , \, \mathrm{rk}(\bS) \leq p}{\arg\min} \,\, 
    \left\{ 
    \sum\limits_{\mu=1}^{n} 
    \left(y_{\mu}-{\rm Tr}\!\left[\bS \, \bG_\mu\right]\right)^{2} 
    +  d\tl\Tr(\bS)
    \right\},
\end{equation}
where $\tilde\lambda=\lambda\sqrt{p/d}$. 
This formulation makes explicit that the width of the neural network acts as a rank constraint on the learned positive semidefinite matrix, while the weight decay in the original parametrization becomes a trace regularization in the matrix formulation, mapping the original problem to matrix compressed sensing with nuclear norm regularization \citep{fazel2008compressed} and a non-convex rank constraint.
Since the network is quadratic in the input, it can only learn target functions that lie in the span of quadratic features. 
We therefore consider labels generated by a centered quadratic model,
\begin{equation}
\label{eq:def:target}
    y_{\mu} 
    = \Tr[\bS_\star (\bx_\mu\bx_\mu^{\top} - \bI_{d})/\sqrt{d}]+ \sqrt{\Delta}\,\xi_\mu 
    \, .
\end{equation}
Here $\{\bx_\mu\}_{\mu=1}^n\overset{\mathrm{i.i.d.}}{\sim}\mathcal{N}(0,\bI_d)$ denotes the input samples, and 
$\{\xi_\mu\}_{\mu=1}^n\overset{\mathrm{i.i.d.}}{\sim}\mathcal{N}(0,1)$ is independent Gaussian label noise, with variance parameter $\Delta\geq0$. 
The quadratic form represents the structured component of the target, which is in the model class of the network and can therefore be learned from data. 
The noise term represents an unstructured component of the labels: it carries no predictive information about fresh inputs.
We further assume that the target matrix $\bS_{\star}$ is symmetric and has power-law decaying spectrum, i.e. that
it has eigenvalues $\{\sqrt{d}\, i^{-\gamma} / \sqrt{\zeta(2\gamma)} \}_{i=1}^d$, where $\zeta$ is Riemann's zeta function and we impose $\gamma > 1/2$ to ensure square summability. The normalization is such that $\Tr( \bS_\star^2 ) / d  = 1 + o_d(1)$. 

Our main goal in this paper is to study the excess test error
\begin{align}\label{eq:def:excess_test error}
    \caE(\bhW) = \frac{1}{2} \mathbb{E}_{\bx, y}\left[ (y - f_p(\bx;\bhW))^{2}\right] - \frac{\Delta}{2}\,,
\end{align}
of the minimizers $\bhW$ of \eqref{eq:def:erm},
where $(\bx, y)$ is a new data pair with $\bx \sim \mathcal{N}(0, \bI_d)$ and $y$ given by \eqref{eq:def:target}. 
We want to characterize its scaling behavior as a function of the number of trainable parameters $N_{\rm param} = dp$, the number of available training samples $n$ and the regularization~$\tl$.
We will be interested in general scaling regimes and for this purpose we define the following high-dimensional limit 
\begin{equation}
\label{eq:limit}
d \gg 1,\qquad
p = \Theta(d^\rho),\qquad  n = \Theta(d^{\alpha}), \qquad \tl = \Theta(d^{\ell}) \qquad {\rm with} \qquad \alpha, \rho > 0 \, , 
\end{equation} 
where $d$ will be taken large enough, but still finite.

\begin{figure}[t]
\centering
\includegraphics[width=\linewidth]{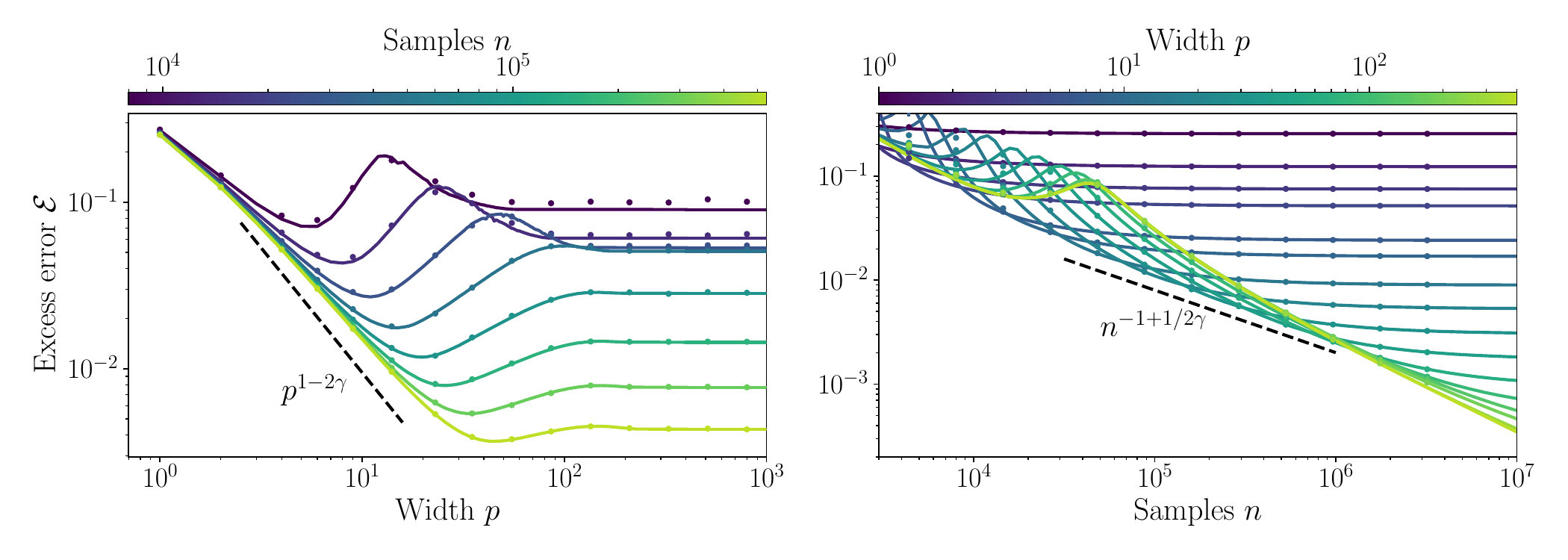}
\caption{
    Excess test error vs  width $p$ (left) and number of samples $n$ (right), for $d = 400$, $\gamma = 1.25$ and $\tl =  \sqrt{n/100d^2}$. 
    Lines are theoretical predictions from Result \ref{res:se}, dots are numerical simulations by LBFGS on \eqref{eq:def:erm}, see Appendix \ref{app:numerics}.
    As a function of the width, we highlight the low-$p$ data-dependent decay of the error with exponent $1-2\gamma$ (dashed black).
    As a function of the number of samples, we highlight that networks at the optimal width, given in Eq.~\eqref{eq:rho_opt_small_reg}, (envelope of the curves) coincide with the Bayes optimal rate $-1+\frac{1}{2\gamma}$ from \cite{defilippis2026scaling} (dashed black).
}
\label{fig:fig1}
\end{figure}

\paragraph{Summary of main results.}
Our results provide a predictive theory of how model size acts as a control parameter in a feature-learning model.  

\begin{itemize}
    \item We derive a characterization of global
    minimizers of \eqref{eq:def:erm} as a function of the network width
    $p$, the number of samples $n$, the effective regularization
    $\tl=\sqrt{\frac{p}{d}}\,\lambda$, and the spectral structure of the target,
    parametrized by $\gamma$; see Result~\ref{res:se}.  The equations
    predict the excess test error and reveal how explicit
    regularization and the implicit rank constraint induced by finite
    width jointly shape the learned predictor.

    \item We obtain explicit scaling laws for the excess test error as a
    function of $p,n,\tl$; see Results~\ref{res:under-reg} and
    \ref{res:over-reg}.  These laws show that width is not only a
    capacity parameter: it also acts as an implicit regularizer.  At
    small width, the error is dominated by the unresolved tail of the
    power-law target.  At larger width, finite-sample effects enter, and
    increasing the number of parameters can improve generalization,
    leave it essentially unchanged, or eventually worsen it by fitting
    noise-dominated directions.\looseness=-1

    \item We characterize the optimal choices of width and
    regularization.  Under mild assumptions, either optimally tuning the
    explicit regularization or selecting the optimal finite width
    achieves the Bayes-optimal excess-error rate.  Thus the
    rank constraint induced by width can provide an implicit
    regularization mechanism that is, at the level of scaling exponents,
    as effective as an explicit optimal one.\looseness=-1
\end{itemize}

Figure~\ref{fig:fig1} summarizes these phenomena.  The left panel shows
the predicted and observed excess test error as a function of width,
including the low-width decay controlled by the power-law exponent
$\gamma$ and the emergence of an optimal width.  The right panel shows
the corresponding sample-wise scaling: optimizing over the width traces
the envelope of the curves and reaches the Bayes-optimal rate.  The
agreement between the state-evolution prediction and numerical
optimization supports the use of Result~\ref{res:se} as a quantitative
description of the ERM.

\paragraph{Further related works.} Our analysis leverages tools from the theory of Approximate Message Passing algorithms (AMP)
\citep{bayati2011lasso,donoho13,berthierStateEvolutionApproximate2020,gerbelot2023graph,vilucchio2025asymptotics},
which have been broadly applied to the study of learning problems in the high-dimensional limit
\citep{donoho2016high,rangan2016fixed}.
In this work, we use the theory of AMP in a non-asymptotic regime as a heuristic extension of its standard asymptotic setting. 
This is in line with a common route in high-dimensional statistics, where heuristic predictions \citep{bordelon2020spectrum,cui2021generalization,simon2023eigenlearning} often precede and guide later rigorous analyses \citep{cheng2024dimension,misiakiewicz2024non,defilippis2024dimension}. 
We view the rigorous justification of this non-asymptotic extension as a challenging mathematical problem in its own right, but distinct from the main goal of the present work.

Quadratic neural networks, their landscape, and sharp generalization properties were studied extensively in the full-rank case, $p\ge d$: \cite{pmlr-v80-du18a,venturi19,soltanolkotabi2018theoretical, gamarnikStationaryPointsShallow2020,martin2024impact,arjevani2025geometry,erba2025nuclear}.
In particular, their scaling laws were derived from the non-asymptotic AMP and for power-law targets in \cite{defilippis2026scaling}, still in the regime $p\ge d$, and extended to more general even activation functions in \cite{defilippis2026optimal}.

The works \cite{ren2025emergence,arous2025learning} study closely
related hierarchical multi-index models and width-constrained quadratic
networks from a dynamical perspective.  Their scaling laws describe
idealized continuous-time population or online trajectories, rather than
the discretized SGD dynamics or the finite-sample ERM considered here. While these works provide an important dynamical account of how
width affects feature learning, they do not address the finite-dataset
generalization problem studied in this paper, including train--test
gaps, interpolation, and width-dependent implicit regularization.

The rank/width constraint regime, $1 \ll p \le d$, was also studied in 
\cite{martin2026high}, where the authors consider the long-time behavior of gradient descent 
    $\bW^{t+1} - \bW^t = \eta \nabla \mathcal{L}(\bW^t)$
in the limit of $\eta \to 0$, $t \to \infty$ with initialization $\bw^{t=0}_i \sim \mathcal{N}(0,\bI_d)$ for $1\leq i \leq p$. The authors obtain an asymptotic characterization of the stationary states of the dynamics in the limit of large $d$ and quadratic sample complexity $n\propto d^2$, which is consistent with our results in their asymptotic version.

Finally, we note that quadratic neural networks are closely related to attention mechanisms, since both operate through bilinear projections of the inputs, as discussed in Appendix \ref{app:attention}. This connection extends beyond structure to their high-dimensional analysis. Recent work \citep{boncoraglio2026singleheadattentionhighdimensions} characterizes the learning curves of a single attention head, showing that the scaling laws obtained in \citep{defilippis2026scaling} for the quadratic model also arise for key and query width $p\ge d$. We expect the finite-width analysis developed here to extend similarly to attention heads with arbitrary width, as discussed in Appendix \ref{app:attention}.

\paragraph{Notations.}
We say that two functions $f(d) \sim g(d)$ if $\lim_{d\to +\infty} f(d)/g(d) = 1$, and that $f(d) = \Theta(g(d))$ if $c g(d) \leq f(d) \leq C g(d)$ for some constants $c,C>0$. Finally, we denote with $\overset{d}{\sim}$ convergence in distribution. 
$\text{GOE}(d)$ denotes the ensemble of symmetric $d \times d$ matrices with Gaussian entries with mean zero and variance $1/d$ ($2/d$ on the diagonal).

\section{Width-wise and sample-wise scaling laws}

Our first result is the analytic characterization of the global minimizer $\bhW$ of \eqref{eq:def:erm} through its equivalent characterization in terms of $\bhS = \bhW^\top \bhW / \sqrt{pd}$ and its associated minimization problem \eqref{eq:def:erm_equiv_intro}.
\begin{result}[Analytic properties of the ERM]
\label{res:se}
Consider any global minimum $\bhW$ of \eqref{eq:def:erm} where the labels are generated from the model in \eqref{eq:def:target} with $\Delta \geq 0$, and define $\bhS = \bhW^\top \bhW / \sqrt{pd}$. Assume the target matrix $\bS_\star \in \bbR^{d \times d}$ has eigenvalues $\{\sqrt{d} i^{-\gamma}/ \sqrt{\zeta(2\gamma)}\}_{i=1}^d$, where $\zeta$ is Riemann's zeta function and $\gamma > 1/2$.
Then for $n, p, d \gg 1$
we have
\begin{equation}\label{eq:SE_test error}
    \bhS \overset{d}{\sim} \bhS_{\delta,\tl\epsilon} = (\bS_\star + \delta \bZ - \tl \epsilon \bI_d)^+_{(p)}
    \, , \quad
    \caE(\bhW) \sim \frac{2n}{d^2}\delta^2 - \frac{\Delta}{2} 
    \, , \quad
    \caL(\bhW) \sim
    \frac{\delta^2}{4\epsilon^2} + \frac{\tl}{d} \Tr(\bhS_{\delta,\tl\epsilon})
    \, ,
\end{equation}
where $\bA^+_{(p)}$ denotes the projection of the matrix $\bA$ onto PSD, rank-$p$ matrices\footnote{ 
$\bA^+_{(p)}$ denotes the matrix $\bA$, with the largest $p$ eigenvalues $\nu_i$ mapped to $\max(\nu_i, 0)$, and the rest set to 0.}, and $\bZ \sim \text{GOE}(d)$. 
Here $(\delta,\epsilon) \in \bbR^2$ are functions of $(n,d,p,\lambda,\Delta,\gamma)$, found as a solution of the system
\begin{equation}
    \begin{cases}
    \begin{aligned}
        \displaystyle
        &\frac{4n}{d^2}\,\delta - \frac{\delta}{\epsilon}
        =
        \partial_1 J_p(\delta,\tl\epsilon),
        \\[0.4em]
        \displaystyle
        &1 + \frac{\Delta}{2}
        + \frac{2n}{d^2}\,\delta^2
        - \frac{\delta^2}{\epsilon}
        =
        \left(1-\tl\epsilon\,\del_2\right)
        J_p(\delta,\tl\epsilon),
    \end{aligned}
    \end{cases}
    \text{with}\quad
    J_p(\delta,\tl\epsilon) = \frac{1}{d} \Tr\left( \bhS_{\delta,\tl\epsilon}^2 \right)
    \, ,
\label{eq:SE_ERM}
\end{equation}
where $\partial_1,\partial_2$ indicate the partial derivative with respect to the first and second arguments of $J_p$ respectively.
\end{result}

Result~\ref{res:se} gives a closed state-evolution prediction for the test error, train error, and learned weights of the ERM.  Given $n,p,d,\tl$ and the spectrum of $\bS_\star$, one solves \eqref{eq:SE_ERM} numerically and plugs the solution into \eqref{eq:SE_test error}; see Appendix~\ref{app:numerics}. The resulting picture is that the learned representation $\bhS$ is a noisy, spectrally sparsified version of the target $\bS_\star$: explicit regularization through $\tl$ and the implicit rank constraint induced by the width $p$ both bias the network towards the strongest target directions. We use these non-asymptotic equations as predictions across the scaling regimes \eqref{eq:limit}, and validate them extensively by training the original quadratic network with gradient-based optimization.  The agreement is excellent, already for moderate dimensions $d=400$ and small width $p=1$ (Figure~\ref{fig:fig1}).

The derivation in Appendix~\ref{app:derivation_SE} follows the standard AMP/statistical-mechanics route and yields the \emph{replica-symmetric} state-evolution prediction \cite{mezard1988spin,mezard2009information,advani2013statistical,zdeborova2016statistical}. The canonical computation is formulated in the regime $n=\Theta(d^2)$, $p \geq d$, $\tl=\Theta(1)$, assuming a limiting spectral description for $\bS_\star$ \cite{erba2025nuclear}. Extending the resulting equations to the power-law targets and to the regimes \eqref{eq:limit} should be viewed as a non-asymptotic extension, in the spirit of \cite{defilippis2026scaling} (see \cite{dietrich2001statistical} for early reports). This could be formalized, for instance, by first keeping a finite number $m^\star$ of signal directions and then passing to the power-law limit, as in \cite{defilippis2026optimal}. The substantive issue is instead the rank constraint for $p<d$ in the equivalent formulation \eqref{eq:def:erm_equiv_intro}.
This is what makes the width an implicit regularizer, but it also makes the matrix problem non-convex, so the convex proof strategy of \cite{erba2025nuclear} (covering the case $p \geq d$) does not transfer directly.  Thus Result~\ref{res:se} should be read as the replica-symmetric prediction for this rank-constrained problem, and its rigorous proof remains an open challenge.
Numerically, however, 
Result~\ref{res:se} matches the error of gradient descent at convergence with high accuracy.  Moreover, in the asymptotic regime, \eqref{eq:SE_ERM} coincides with the equations derived in \cite{martin2026high} for the long-time behavior of gradient flow in the same model, providing an additional consistency check.

The main consequence of Result~\ref{res:se} is a characterization of the scaling of the excess test error \eqref{eq:def:excess_test error} with the width $p$, sample size $n$, and effective regularization $\tl$, as $d\to\infty$ with $\Delta=O(1)$.  
We express these laws through the exponents $\alpha,\rho,\ell>0$ in \eqref{eq:limit}, fixing $\gamma>1/2$ and $\Delta>0$ (see Appendix \ref{app:noiseless} for the noiseless case $\Delta = 0$).  
The resulting phase diagram is summarized
in Figure~\ref{fig:fig2}; the derivation, including asymptotic
constants depending on $\gamma$ and $\Delta$, is given in
Appendix~\ref{app:derivation_scaling}.  
We distinguish the
under-regularized regime $\ell<\alpha/2-1$ with $\alpha>1$, the
over-regularized regime $\alpha/2-1<\ell<\alpha-3/2$ with $\alpha>\ell+3/2$,
and the rank-collapse regime $\alpha<\max(1,\ell+3/2)$.
The case $\rho=1$ is delicate, and our derivation of the scalings does not cover it: we recover it by taking the limits $\rho\to1^\pm$.

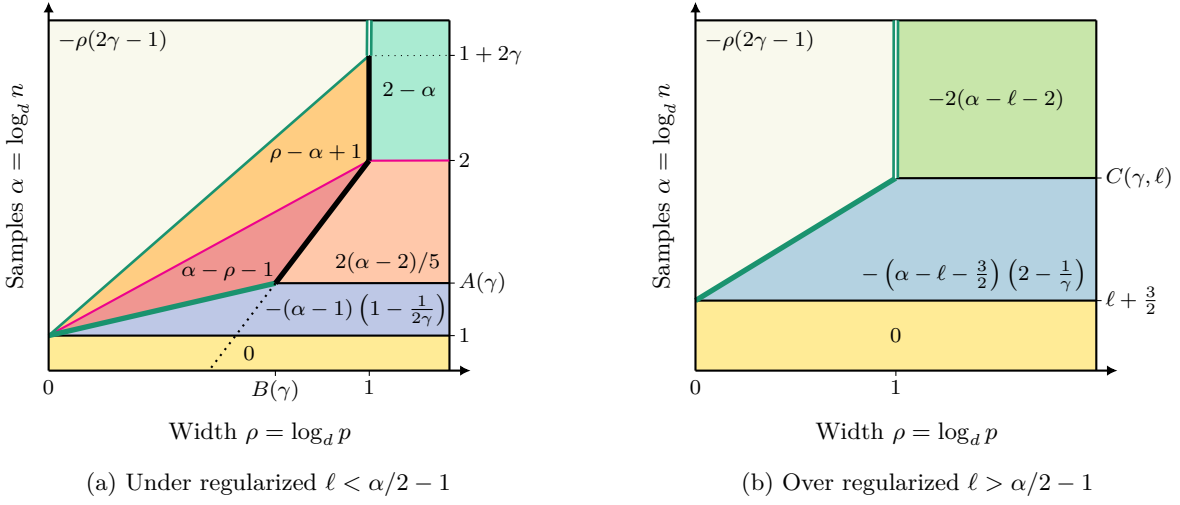
\begin{figure}[t]
\centering
\begin{subfigure}[t]{0.48\linewidth}
    \centering
    \begin{tikzpicture}[scale=13.96]

\tikzstyle{every node}=[font=\small ]




\definecolor{bulk}{RGB}{210,210,210}

\definecolor{phaseI}{RGB}{255,235,150}
\definecolor{phaseII}{RGB}{180,210,225}
\definecolor{phaseIII}{RGB}{200,230,170}
\definecolor{phaseIV}{RGB}{190,200,230}
\definecolor{phaseV}{RGB}{255,200,170}
\definecolor{phaseVI}{RGB}{170,235,210}

\definecolor{phaseLow}{RGB}{249,249,237}
\definecolor{phaseOverfit}{RGB}{255,205,130}
\definecolor{phaseDecay}{RGB}{240,150,145}

\definecolor{mygreen}{RGB}{26, 147, 111}    

\def\Ox{0.08}
\def\Oy{0.09}
\def\AXx{0.40}
\def\AXy{0.35}
\def\height{0.95*\AXy}
\def\width{0.95*\AXx}

\def\overparametrixed{0.8*\width}
\def\plateau{0.4*\width}

\def\linear{0.1*\height}
\def\peak{0.25*\height}
\def\quadratic{0.6*\height}
\def\extreme{0.9*\height}

\coordinate (origin) at (\Ox, \Oy);
\coordinate (LYtop) at (\Ox, \Oy + \height);
\coordinate (RYtop) at (\Ox + \width, \Oy + \height);
\coordinate (RYbottom) at (\Ox + \width, \Oy);

\coordinate (LY1) at (\Ox, \Oy + \linear);
\coordinate (LY2) at (\Ox, \Oy + \peak);
\coordinate (LY3) at (\Ox, \Oy + \quadratic);
\coordinate (LY4) at (\Ox, \Oy + \extreme);

\coordinate (RY1) at (\Ox + \width, \Oy + \linear);
\coordinate (RY2) at (\Ox + \width, \Oy + \peak);
\coordinate (RY3) at (\Ox + \width, \Oy + \quadratic);
\coordinate (RY4) at (\Ox + \width, \Oy + \extreme);

\coordinate (BX1) at (\Ox + \plateau, \Oy);
\coordinate (BX2) at (\Ox + \overparametrixed, \Oy);

\coordinate (TX2) at (\Ox + \overparametrixed, \Oy + \height);

\path[name path=OVER] (BX2) -- (TX2);

\path[name path=LINEAR]     (LY1) -- (RY1);
\path[name path=PEAK]       (LY2) -- (RY2);
\path[name path=QUADRATIC]  (LY3) -- (RY3);
\path[name path=EXTREME]    (LY4) -- (RY4);

\path[name intersections={of=OVER and EXTREME, by=O2}];
\path[name intersections={of=OVER and QUADRATIC, by=O1}];

\path[name path=OBLIQUE]    (BX1) -- (O1);
\path[name intersections={of=OBLIQUE and PEAK, by=P}];
\path[name intersections={of=OBLIQUE and LINEAR, by=PN}];

\path[name path=PLATEAU]     (LY1) -- (O2);
\path[name intersections={of=PLATEAU and QUADRATIC, by=T2}];
\path[name intersections={of=PLATEAU and PEAK, by=T1}];

\coordinate (PX) at (P |- origin);

\path[name intersections={of=OVER and EXTREME, by=O2}];

\fill[phaseI!100]
  (origin) -- (RYbottom) -- (RY1) -- (LY1) -- cycle;
\node at (barycentric cs:origin=1,RYbottom=1,RY1=1,LY1=1) {\footnotesize $0$};

\fill[phaseIV!100]
  (LY1) -- (RY1)  -- (RY2) -- (P) -- cycle;
\node[anchor=south east, xshift=3pt] (r) at (RY1) {\footnotesize $-(\alpha -1 )\left(1-\frac{1}{2\gamma}\right)$};

\fill[phaseV!100]
  (P) -- (RY2) -- (RY3)  -- (O1) -- cycle;
\node[anchor=south east] (r) at (RY2) {\footnotesize $2(\alpha - 2)/5$};

\fill[phaseVI!100]
  (O1) -- (RY3) -- (RYtop)  -- (TX2) -- cycle;
\node at ($(O1)!0.5!(RYtop)$) {\footnotesize $2-\alpha$};

\fill[phaseLow!100]
  (LY1) -- (O2) -- (TX2)  -- (LYtop) -- cycle;
\node[anchor=north west] (r) at (LYtop) {\footnotesize $-\rho(2\gamma-1)$};

\fill[phaseDecay!100]
  (LY1) -- (P) -- (O1) -- cycle;
\node[anchor=south east, xshift=3pt, yshift=-2.5pt] (r) at (P) {\footnotesize $\alpha-\rho - 1$};

\fill[phaseOverfit!100]
  (LY1) -- (O1) -- (O2) -- cycle;
\node[anchor=south east, xshift=1pt, yshift=-4pt] (r) at (O1) {\footnotesize $\rho-\alpha + 1$};


\draw[black, thick] (LYtop) -- (RYtop);
\draw[black, thick] (RYtop) -- (RYbottom);

\draw[black, thick] (LY1) -- (RY1);
\draw[black, thick] (P) -- (RY2);


\draw[magenta, thick] (LY1) -- (O1);
\draw[magenta, thick] (O1) -- (RY3);

\draw[mygreen, thick, line width = 1pt] (LY1) -- (O2);

\draw[mygreen, thick, line width = 2pt] (LY1) -- (P);
\draw[black, thick, line width = 2pt] (P) -- (O1);
\draw[black, thick, line width = 2pt] (O1) -- (O2);
\draw[mygreen, thick, line width = 1pt, double] (O2) -- (TX2);

\draw[black, thick, dotted] (P) -- (BX1);




\draw[black, dotted] (O2) -- (RY4);



\draw[->,thick] (origin) -- (\Ox + \AXx,\Oy);
\draw[->,thick] (\Ox,\Oy) -- (\Ox,\Oy + \AXy);

\node[anchor=north] (r) at (\Ox + \AXx/2,\Oy-0.04) {Width $\rho = \log_d p$};
\node[anchor=south, rotate=90] (r) at (\Ox-0.01,\Oy + \AXy/2) {Samples $\alpha = \log_d n$};

\node[anchor=north] (r) at (origin) {\footnotesize $0$};

\draw (BX2) -- ($(BX2)+(0,-0.005)$);
\node[anchor=north] (r) at (BX2) {\footnotesize $1$};

\draw (PX) -- ($(PX)+(0,-0.005)$);
\node[anchor=north] (r) at (PX) {\footnotesize $B(\gamma)$};

\draw (RY4) -- ($(RY4)+(0.005,0)$);
\node[anchor=west, rotate=0] at (RY4) {\footnotesize $1+2\gamma$};

\draw (RY3) -- ($(RY3)+(0.005,0)$);
\node[anchor=west, rotate=0] (r) at (RY3) {\footnotesize $2$};

\draw (RY2) -- ($(RY2)+(0.005,0)$);
\node[anchor=west, rotate=0] (r) at (RY2) {\footnotesize $A(\gamma)$};

\draw (RY1) -- ($(RY1)+(0.005,0)$);
\node[rotate=0, anchor=west] (r) at (RY1) {\footnotesize $1$};







\end{tikzpicture}
    \caption{Under regularized $\ell < \alpha/2-1$}
\end{subfigure}
\hfill
\begin{subfigure}[t]{0.48\linewidth}
    \centering
    \begin{tikzpicture}[scale=13.96]

\tikzstyle{every node}=[font=\small]



\definecolor{bulk}{RGB}{210,210,210}

\definecolor{phaseI}{RGB}{255,235,150}
\definecolor{phaseII}{RGB}{180,210,225}
\definecolor{phaseIII}{RGB}{200,230,170}
\definecolor{phaseIV}{RGB}{190,200,230}
\definecolor{phaseV}{RGB}{255,200,170}
\definecolor{phaseVI}{RGB}{170,235,210}

\definecolor{phaseLow}{RGB}{249,249,237}

\definecolor{mygreen}{RGB}{26, 147, 111}    

\def\Ox{0.08}
\def\Oy{0.09}
\def\AXx{0.40}
\def\AXy{0.35}
\def\height{0.95*\AXy}
\def\width{0.95*\AXx}

\def\overparametrixed{0.5*\width}
\def\plateau{0.4*\width}

\def\linear{0.2*\height}
\def\peak{0.25*\height}
\def\quadratic{0.55*\height}
\def\extreme{0.9*\height}

\coordinate (origin) at (\Ox, \Oy);
\coordinate (LYtop) at (\Ox, \Oy + \height);
\coordinate (RYtop) at (\Ox + \width, \Oy + \height);
\coordinate (RYbottom) at (\Ox + \width, \Oy);

\coordinate (LY1) at (\Ox, \Oy + \linear);
\coordinate (LY2) at (\Ox, \Oy + \peak);
\coordinate (LY3) at (\Ox, \Oy + \quadratic);
\coordinate (LY4) at (\Ox, \Oy + \extreme);

\coordinate (RY1) at (\Ox + \width, \Oy + \linear);
\coordinate (RY2) at (\Ox + \width, \Oy + \peak);
\coordinate (RY3) at (\Ox + \width, \Oy + \quadratic);
\coordinate (RY4) at (\Ox + \width, \Oy + \extreme);

\coordinate (BX1) at (\Ox + \plateau, \Oy);
\coordinate (BX2) at (\Ox + \overparametrixed, \Oy);

\coordinate (TX2) at (\Ox + \overparametrixed, \Oy + \height);

\path[name path=OVER] (BX2) -- (TX2);

\path[name path=LINEAR]     (LY1) -- (RY1);
\path[name path=PEAK]       (LY2) -- (RY2);
\path[name path=QUADRATIC]  (LY3) -- (RY3);
\path[name path=EXTREME]    (LY4) -- (RY4);

\path[name intersections={of=OVER and EXTREME, by=O2}];
\path[name intersections={of=OVER and QUADRATIC, by=O1}];

\path[name path=OBLIQUE]    (BX1) -- (O1);
\path[name intersections={of=OBLIQUE and PEAK, by=P}];

\path[name path=PLATEAU]     (LY1) -- (O2);
\path[name intersections={of=PLATEAU and QUADRATIC, by=T2}];
\path[name intersections={of=PLATEAU and PEAK, by=T1}];

\coordinate (PX) at (P |- origin);

\fill[phaseII!100]
  (LY1) -- (O1)  -- (RY3) -- (RYbottom) -- cycle;
\node[anchor=south east] (r) at (RY1) {\footnotesize $-\left( \alpha-\ell - \frac32 \right)\left(2-\frac1\gamma\right)$};

\fill[phaseIII!100]
  (O1) -- (RY3) -- (RYtop)  -- (TX2) -- cycle;
\node at ($(O1)!0.5!(RYtop)$) {\footnotesize $-2(\alpha-\ell-2)$};

\fill[phaseLow!100]
  (LY1) -- (O1) -- (TX2)  -- (LYtop) -- cycle;
\node[anchor=north west] (r) at (LYtop) {\footnotesize $-\rho(2\gamma-1)$};

\fill[phaseI!100]
  (origin) -- (RYbottom) -- (RY1) -- (LY1) -- cycle;
\node at (barycentric cs:origin=1,RYbottom=1,RY1=1,LY1=1) {\footnotesize $0$};

\draw[black, thick] (LYtop) -- (RYtop);
\draw[black, thick] (RYtop) -- (RYbottom);

\draw[black, thick] (O1) -- (RY3);
\draw[black, thick] (LY1) -- (RY1);

\draw[mygreen, line width=2pt] (O1) -- (LY1);
\draw[mygreen, line width=1pt, double] (O1) -- (TX2);


\draw[->,thick] (origin) -- (\Ox + \AXx,\Oy);
\draw[->,thick] (\Ox,\Oy) -- (\Ox,\Oy + \AXy);

\node[anchor=north] (r) at (\Ox + \AXx/2,\Oy-0.04) {Width $\rho = \log_d p$};
\node[anchor=south, rotate=90] (r) at (\Ox-0.01,\Oy + \AXy/2) {Samples $\alpha = \log_d n$};

\node[anchor=north] (r) at (origin) {\footnotesize $0$};

\draw (BX2) -- ($(BX2)+(0,-0.005)$);
\node[anchor=north] (r) at (BX2) {\footnotesize $1$};

\draw (RY3) -- ($(RY3)+(0.005,0)$);
\node[anchor=west, rotate=0] (r) at (RY3) {\footnotesize $C(\gamma, \ell)$};

\draw (RY1) -- ($(RY1)+(0.005,0)$);
\node[rotate=0, anchor=west] (r) at (RY1) {\footnotesize $\ell+\frac{3}{2}$};







\end{tikzpicture}
    \caption{Over regularized $\ell>\alpha/2-1$}
\end{subfigure}
\caption{
    Scaling laws as a function of the number of training samples $n = \Theta(d^\alpha)$, of the width of the network $p = \Theta(d^\rho)$ and of the decay of the target weights $\gamma$ for the cases of low regularization (left) and large regularization (right). 
    In each phase, we write the scaling exponent $\beta$ of the test error $\caE= \Theta(d^{\beta})$. 
    With green lines, we highlight the optimal width scaling (where two values of $\rho$ are highlighted at same $\alpha$, the optimal width can be either of both based on $\gamma$ and $\Delta$, see \eqref{eq:rho_opt_small_reg}).
    With thick lines, we highlight the effective target's rank $\rho_{\rm eff}$, to the right of which the error scaling is the same as for full-width networks.
    With double line, we mark a transition across which the scaling exponent of the test error is discontinuous.
    With red line, we highlight the interpolation peak, below which the training set can be fitted exactly.
    We defined $A(\gamma) = (18\gamma-5)/(14\gamma-5)$, $B(\gamma) = (8\gamma-2)/(14\gamma-5)$ and $C(\gamma, \ell) = \ell + \gamma + 3/2$.
    All phases to the left of the bold or dashed line are newly derived scalings, while the other phases can be derived from the $p \geq d$ theory in \cite{defilippis2026scaling}.
}
\label{fig:fig2}
\end{figure}

\paragraph{Under-regularized regime.} If $\alpha > 1$ (enough samples) and $\ell < \alpha/2 - 1$ (low regularization), we obtain the following excess test error scaling behavior.
\begin{result}[Scaling law for low regularization]
    \label{res:under-reg}
    In the setting of Result \ref{res:se}, when $\alpha > 1$ and $\ell < \alpha/2 - 1$ we have $\caE= \Theta(d^\beta)$, where
    \begin{equation}
        \beta = 
        \begin{cases}
            - (2\gamma-1) \rho
            &\text{if}\quad 
            \rho < \min\left\{ (\alpha - 1) / (2\gamma), 1\right\}
            \\
            \rho + 1 - \alpha
            &\text{if}\quad 
            (\alpha - 1) / (2\gamma) < \rho < \min(1,\alpha - 1)
            \mathand 
            \alpha < 1+2\gamma
            \\
            \alpha - 1 - \rho
            &\text{if}\quad 
            \min(1,\alpha - 1) < \rho < \rho_{\rm eff, 1}(\alpha, \gamma)
            \mathand 
            \alpha < 2 
            \\
            \phi_1(\alpha, \gamma)
            &\text{if}\quad 
            \rho > \rho_{\rm eff, 1}(\alpha, \gamma)
        \end{cases}
    \end{equation}
    where, calling $A(\gamma) = (18\gamma-5)/(14\gamma-5) \in [9/7, 2]$ (recall $\gamma >1/2$), we defined
    \begin{equation}
        \phi_1(\alpha, \gamma) = 
        \begin{cases}
            -(\alpha-1) \left(1-\frac{1}{2\gamma}\right)
            &\text{if}\quad \alpha < A(\gamma)
            \\
            2(\alpha-2) / 5
            &\text{if}\quad A(\gamma) < \alpha < 2
            \\
            2 - \alpha
            &\text{if}\quad \alpha > 2
        \end{cases}
        \, ,
    \end{equation}
    and
    \begin{equation}
        \rho_{\rm eff, 1}(\alpha, \gamma) =  
        \begin{cases}
            (\alpha - 1) (4\gamma - 1) / (2\gamma)
            &\text{if}\quad \alpha < A(\gamma)
            \\
            (3 \alpha - 1)/5 
            &\text{if}\quad A(\gamma) < \alpha < 2
            \\
            1
            &\text{if}\quad \alpha > 2
        \end{cases} \, .
    \end{equation}
\end{result}
In this regime, the regularization is low enough that its presence does not alter the rates.
The quantity $\rho_{\rm eff, 1}(\alpha, \gamma)$ can be interpreted as an effective rank of the target weights (as seen through the finite sample size $n$).
For $\rho > \rho_{\rm eff, 1}(\alpha, \gamma)$,  the rates coincide with the ones of a full-rank student as in \cite{defilippis2026scaling}.

Let us comment the phases one by one (recalling that in this regime $\alpha > 1$ and $\ell < \alpha/2 - 1$); small arrows after the phase name indicate the monotonicity of the excess test error as a function of the network's width for fixed number of samples). 
We have two phases that are present for all values of sample scaling $\alpha$.
\begin{itemize}
    \item \textbf{Low width $(\searrow)$} with data-dependent exponent $-\rho(2\gamma-1)$. When the width scaling $\rho$ is low enough the ERM weights have $p = \Theta(d^\rho)$ spikes in the spectrum, each correlating with one of the $\Theta(d^\rho)$ dominant eigen-spaces of the target. The excess test error is composed of two terms, an approximation error due to the imperfect spike-to-eigen-space correlation, and a low-rank approximation error due to the $d-p$ un-recovered target eigen-spaces. The low-rank approximation error dominates and determines the asymptotic scaling of the excess test error.
    \item \textbf{Full width $(\rightarrow)$} with $\rho$-independent exponents. For $\rho > \rho_{\rm eff, 1}$, the ERM plateaus to a $\rho$-independent scaling. The value of the plateau depends on the number of samples. If $\alpha < 2$, the plateau showcases a non-trivial behavior first decreasing and then increasing as a function of $\alpha$ related to overfitting of the label noise. For $\alpha > 2$ instead, the plateau decreases as $-\alpha$, as the number of samples is large enough to avoid overfitting. We stress that for $1 < \alpha < A(\gamma)$ and $\rho$ small enough the width constraint is active, but this has no effect at leading order for the error scaling.
\end{itemize}
For $\alpha > 1+2\gamma$ these are the only two phases. As $\alpha$ gets lowered, first getting below $1+2\gamma$ and then below $2$, the other phases enter the phase diagram, providing a richer phenomenology between the low width and the full width phases.
\begin{itemize}
    \item \textbf{Overfitting before interpolation $(\nearrow)$} with $\gamma$-independent exponent $\rho-\alpha+1$.
    The first additional behavior, present as soon as $\alpha < 1+2\gamma$, is a monotone increasing behavior of the excess test error in $\rho$. This is caused by the fact that the ERM weights have strictly less then $p$ outlying eigen-spaces correlating with the target, while the rest of the non-zero eigenvalues are spurious, i.e. their eigenvector do not correlate with the target and they are caused by label noise overfitting. As $\rho$ increases, there are more and more of these spurious eigenvalues coalescing into a bulk, leading to a test error that increases with the width.
    \item \textbf{Decay after interpolation $(\searrow)$} with $\gamma$-independent exponent $\alpha-\rho-1$.
    The second additional behavior, present as soon as $\alpha < 2$, is a further decay. As the width of the network increases, the bulk of spurious eigenvalues shrink, reducing the test error and providing a benefit of overparametrization.
\end{itemize}
Remarkably, in the extremely over-sampled scaling $\alpha > 2\gamma+1$ where only the low width and full width are present, we observe a discontinuous jump in scaling between the two phases at $\rho = 1$: for $\rho\to 1^-$ the error scales as  $\Theta(d^{-2\gamma})$, while for $\rho>1$ scales as $\Theta(d^{2-\alpha})$. 
For $\alpha < 2\gamma + 1$ instead, we observe no 
discontinuity at the boundary $\rho=1$ between width-constrained models and non-width-constrained model.

\paragraph{Over-regularized regime.} If $\alpha >\ell+3/2 $ (enough samples) and $\alpha/2-1 < \ell < \alpha - 3/2$ (large regularization), we obtain the following excess test error scaling behavior.
\begin{result}[Scaling law for large regularization]
    \label{res:over-reg}
    In the setting of Result \ref{res:se}, when $\alpha > 1$ and $\alpha/2-1 < \ell < \alpha - 3/2$ we have $\caE= \Theta(d^\beta)$, where 
    \begin{equation}
        \beta = 
        \begin{cases}
            - (2\gamma-1) \rho 
            &\text{if}\quad 
            \rho < \rho_{\rm eff, 2}(\alpha, \gamma, \ell)
            \\
            \phi_2(\alpha, \gamma)
            &\text{if}\quad \rho > \rho_{\rm eff, 2}(\alpha, \gamma, \ell)
        \end{cases}
    \end{equation}
    where $\phi_2$ is the following width-independent plateau
    \begin{equation}
        \phi_2(\alpha, \gamma) = 
        \begin{cases}
            (\ell + 3/2 - \alpha)(2 - 1/\gamma)
            &\text{if}\quad 
            \alpha < \ell + 3/2 +\gamma
            \\
            2(\ell + 2 - \alpha)
            &\text{if}\quad 
            \alpha > \ell + 3/2 + \gamma
        \end{cases}
        \, ,
    \end{equation}
      and
    \begin{equation}
        \rho_{\rm eff, 2}(\alpha, \gamma, \ell) = 
        \begin{cases}
            (\alpha - \ell - 3/2)/\gamma
            &\text{if}\quad \alpha < \ell + 3/2 +\gamma
            \\
            1
            &\text{if}\quad \alpha > \ell + 3/2 +\gamma
        \end{cases} \, .
    \end{equation}
\end{result}

In this over-regularized regime we observe a similar qualitative behavior as in the under-regularized regime, with two phases present for all $\alpha>1$ at low width and full width with similar qualitative interpretation, separated by the threshold $\rho_{\rm eff, 2}(\alpha, \gamma)$, which again can be interpreted as an effective target width. 
The higher regularization induces sparser weights spectra, and prevents the intermediate-$\rho$ phases related to interpolation observed at low number of samples in the under-regularized regime.
At the boundary between width-constrained network and full width network $\rho = 1$ we again observe a discontinuity in the scaling of the test error for large number of samples $\alpha > \ell + 3/2 + \gamma$.

\paragraph{Rank collapse regime.} If $\alpha < 1$ (extreme under-sampling) or $\ell > \alpha - 3/2$ (extreme over-regularization), the solution of the ERM \eqref{eq:def:erm} sets the weights of the network to zero, as if no samples were provided. The excess test error equals one at leading order, $\caE\sim 1$.

\paragraph{On different kinds of over-parametrization.}
The scaling diagram in Figure \ref{fig:fig2} allows to characterize different kinds of over-parametrization. 
In all phases with $\alpha < \min(2, 1+\rho)$, the network can fit perfectly the training set.
In all phases with $\rho > \rho_{\rm eff,1/2}(\alpha, \gamma)$ (for both under- and over-regularization), the networks properties do not change when the width of the network is enlarged.
In all phases with $\rho > 1$ (or $p\geq d$), the optimization problem \eqref{eq:def:erm} is benign: there are no spurious local minima.

\begin{figure}[t]
\centering
\includegraphics[width=\linewidth]{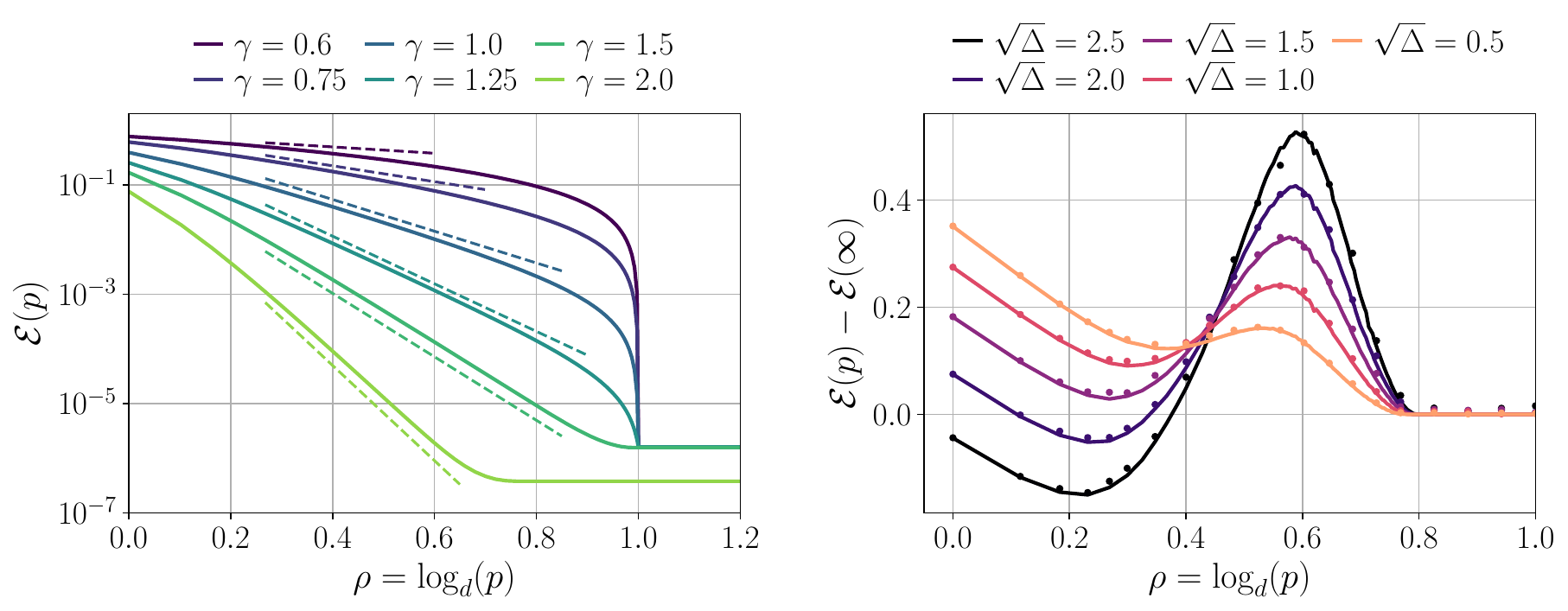}
\caption{
        (Left) Test error scaling behavior in the over-regularized regime for different values of $\gamma$ as a function of the width $p$, for fixed $d=800$, $\sqrt{\Delta}=0.5$, $\alpha=4$ and $\ell= 1.2$. We observe the low-width data-dependent scaling with exponent $1-2\gamma$ (dashed lines, one per value of $\gamma$), as well as the large-width plateau.
    (Right) Dependency of the optimal width on $(\gamma, \Delta)$ in the low-regularization, low-samples regime. We fix $d=400$, $\ell = -0.15$, $\alpha = 1.7$, $\gamma = 0.6$, and vary $\Delta$. We observe that as the label noise decreases the optimal width jumps discontinuously from the position of the local minimizer to the start of the large-width plateau.
    In both plots, lines are theory (Result \ref{res:se}), points numerical experiments (Appendix \ref{app:numerics}).
}
\label{fig:fig3}
\end{figure}

\section{Rates for different regularization schemes}
In this section, we derive scaling laws under optimal choices of regularization strength, network width, and post-training pruning. We treat $\ell_2$ regularization, width constraints, and pruning as three distinct forms of explicit regularization, and ask whether any one of them is inherently superior.
We show that, across all three settings (under mild additional assumptions) the resulting test error can achieve Bayes-optimal scaling rates (as characterized in \cite[Corollary 1]{defilippis2026scaling}). This highlights a nontrivial interplay between these mechanisms: depending on the practical cost of cross-validating each form of regularization, one can recover optimal scaling behavior through multiple, interchangeable routes. For example, optimal post-training pruning requires a single training run on a convex objective, while the other regularization schemes may require multiple ones to perform cross-validation.

\paragraph{Optimal regularization under width constraint.}
From Result \ref{res:under-reg} and \ref{res:over-reg} we can see that the optimal effective regularization $\tl$ is given by the scaling $\ell = \alpha/2 - 1$ for all $\alpha, \rho > 0$. 
Indeed, the over-regularized rates improve monotonously as the regularization decreases, while the under-regularized rates either do not depend on $\tl$ (hence all $\ell < \alpha/2 - 1$ are optimal), or showcase a monotone increasing behavior with the width, which is sub-optimal (overfitting phase).
We stress that the optimal regularization still depends on the width, as
$\lambda_{\rm opt} = \sqrt{d/p} \tl_{\rm opt} = \Theta(d^{(\alpha-\rho -1)/2})$. The associated test error then scales as
\begin{equation}
    \log_d \caE_{\rm opt\, reg} 
    \sim
    \begin{cases}
        - (2\gamma-1) \rho 
        &\text{if}\quad 
        \rho < \rho_{\rm eff, 2}(\alpha, \gamma)
        \\
        (1 - \alpha)(1 - 1/(2\gamma))
        &\text{if}\quad \rho > \rho_{\rm eff, 2}(\alpha, \gamma)
        \mathand 
        \alpha < 1 + 2\gamma
        \\
        2 - \alpha
        &\text{if}\quad \rho > \rho_{\rm eff, 2}(\alpha, \gamma)
        \mathand 
        \alpha > 1 + 2\gamma
    \end{cases} \, ,
\end{equation}
For $\rho > \rho_{\rm eff, 2}(\alpha, \gamma)$, the rates obtained are Bayes optimal, while for $\rho < \rho_{\rm eff, 2}(\alpha, \gamma)$ they are sub-optimal. As long as the network is not width-constrained, optimal regularization can achieve Bayes optimal rates.

\paragraph{Optimal width at fixed regularization.}
We are now interested in computing the optimal width, by which we mean the smallest width achieving the best test error scaling.
In the over-regularized regime
the test error scaling is always monotone decreasing and then plateauing in $\rho$, hence the optimal $\rho$ is the smallest width in the plateauing phase, $\rho^{\rm over \, reg}_{\rm opt \, width} = \rho_{\rm eff,2}(\alpha, \gamma, \ell)$.
The associated test error scaling is given by
\begin{equation}
    \log_d \caE^{\rm over \, reg}_{\rm opt \, width} \sim
    \begin{cases}
        (\ell + 3/2 - \alpha)(2 - 1/\gamma)
        &\text{if}\quad 
        \alpha < \ell + 3/2 +\gamma
        \\
        2(\ell + 2 - \alpha)
        &\text{if}\quad 
        \alpha > \ell + 3/2 +\gamma
    \end{cases}
\end{equation}
which reduces to the Bayes optimal test error for $\ell = \alpha / 2 - 1$, and is suboptimal for larger regularization.

In the under-regularized regime, the test error scaling may be non-monotone. In particular, one needs to compare the error at the end of the low-width decreasing phase (at width $\rho_{\rm eff,1}(\alpha, \gamma)$), where either the large width plateau starts or there is a local minimum of the test error scaling, with the value of the full width plateau.
For $\alpha > 2$ the two coincide. For $\alpha > A(\gamma)$, the local minimum at $(\alpha-1)/(2\gamma)$ has smallest test error scaling.
For $1<\alpha<A(\gamma)$ the scaling of the test error at the local minimum and at the plateau are the same, so optimality is found by comparing the constants.
We find that there exists a function $G(\gamma, \Delta)$ such that
\begin{equation}\label{eq:rho_opt_small_reg}
    \rho^{\rm under \, reg}_{\rm opt \, width} = 
    \begin{cases}
        \min\left\{(\alpha-1)/(2\gamma), 1\right\}
        &\text{if}\,\, 
        \alpha > A(\gamma)
        \,\, \text{or}\,\,  \left(
        1 < \alpha < A(\gamma)
        \,\, \text{and}\,\,  G(\gamma, \Delta) > 0
        \right)
        \\
        \rho_{\rm eff,1}(\alpha, \gamma)
        &\text{if}\,\, 
        1 < \alpha < A(\gamma)
        \,\, \text{and}\,\,  G(\gamma, \Delta) < 0 
    \end{cases}.
\end{equation}
In particular, if $G(\gamma, \Delta) < 0$ the optimal width behaves discontinuously at $A(\gamma)$, jumping from a larger value at small sample scaling to a smaller value at larger sample scaling, revealing an interesting interplay between data structure $\gamma$ and label noise $\Delta$.
To compute explicitly $G(\gamma, \Delta)$ and assess whether it can be negative for $d\gg 1$, we would need to control not only the asymptotic constant of the test error, but also that of the transition regime, which is to our best efforts out of reach. We provide numerical evidence that this can be the case for $d=400$ in Figure \ref{fig:fig3} right.
The associated test error is given by
\begin{equation}
    \log_d \caE^{\rm under \, reg}_{\rm opt \, width}
    \sim
    \begin{cases}
        (1 - \alpha) (1 - 1/(2\gamma))
        &\text{if}\quad \alpha < 2\gamma+1
        \\
        2 - \alpha
        &\text{if}\quad \alpha > 2\gamma+1
    \end{cases},
\end{equation}
which is Bayes optimal.
Thus, as long as the network is not over-regularized, training at optimal width achieves Bayes optimal rates. In particular, optimally tuning the width regularizes the interpolation peak behavior, as we show in Figure \ref{fig:fig1} (right). 

\paragraph{Optimal post-training pruning.}
As a third form of explicit regularization, we consider after-training pruning. This regularization consists in training a network at full width $d$, and later prune it down to $p_{\rm pruning} \ll d$ by substituting the weights with their best rank-$p$ approximation (i.e., putting to zero all but the largest $p_{\rm pruning}$ eigenvalues).
For generic final pruning width, we give the error scaling in Appendix \ref{app:pruning}.
We find, that pruning to the optimal width scaling $\rho_{\rm  opt \, pruning}$ achieves Bayes optimal rates with
\begin{equation}
\rho_{\rm  opt \, pruning}=
\min\left\{(\alpha-1)/(2\gamma), 1\right\}
\mathif \ell<\alpha/2-1 .
\end{equation}
We remark that outside the region $\ell<\alpha/2-1$ and $1<\alpha<1+2\gamma$ pruning does not help, because there is no overfitting. Moreover, $\rho_{\rm  opt \, pruning} = \rho^{\rm under \, reg}_{\rm opt \, width}$ in most of the parameter space, with the only possible exclusion of the region $G(\gamma, \Delta) < 0$.

\section{Discussion and limitations}
We studied the joint scaling of generalization with sample size, width,
and regularization in a finite-width quadratic feature-learning model.
Our analysis reveals a rich phase diagram, with distinct scaling
regimes, interpolation effects, plateaus, and non-monotone dependence on
width.  The main message is that width is not merely a capacity
parameter: in this model, limited width acts as an implicit regularizer,
selecting which directions of the target spectrum are learned and which
are suppressed.  This allows us to characterize the optimal width and to
compare explicit regularization by weight decay with implicit
regularization by architectural width restriction or pruning.  In the
appropriate regimes, optimally tuning either regularization or width
achieves optimal rates.

The main limitation of our approach is the price paid for mathematical tractability.  We
consider a deliberately stylized setting: a shallow quadratic network,
Gaussian inputs, and a structured spectral teacher.  We therefore do not
claim that the precise exponents derived here transfer directly to
realistic deep architectures.  Rather, the value of the model is its
solvability: it provides a controlled setting in which the roles of data
availability, model size, target structure, and regularization can be
isolated and understood explicitly.

Among the open problems, a natural mathematical direction is to turn the
state-evolution prediction of Result~\ref{res:se} into a rigorous
theorem.  This is not a mere technical gap.  Making such finite-size
state-evolution predictions fully rigorous would already require
significant progress in the control of AMP beyond standard asymptotic
settings.  In addition, the rank constraint induced by finite width makes
the equivalent optimization problem non-convex, placing it beyond the
existing convex AMP proof techniques. This therefore requires significant progress
in the rigorous theory of AMP/state evolution for non-convex
width-constrained problems.  Another important direction is to analyze
gradient descent directly and prove that it reaches the same macroscopic
solution as the ERM, as our numerical experiments very strongly suggest, even in this non-convex setting.

\section*{Acknowledgements}
We thank Simon Martin and Blake Bordelon for useful discussions.
We also acknowledge funding from the Swiss National Science Foundation grants SNSF SMArtNet (grant number 212049), OperaGOST (grant number 200021 200390), and DSGIANGO (grant number 225837). This work was supported by the Simons Collaboration on the Physics of Learning and Neural Computation via the Simons Foundation grant ($\#1257412$ (FK) and $\#1257413$ (LZ)).
\bibliography{main.bib}

@book{dietrich2001statistical,
  title={Statistical Mechanics of Neural Networks: Enhancement by Weighting of Examples},
  author={Dietrich, Rainer},
  year={2001},
  publisher={Citeseer}
}

@inproceedings{xu2025fundamental,
  title={Fundamental Limits of Matrix Sensing: Exact Asymptotics, Universality, and Applications},
  author={Xu, Yizhou and Maillard, Antoine and Zdeborov{\'a}, Lenka and Krzakala, Florent},
  booktitle={The Thirty Eighth Annual Conference on Learning Theory},
  pages={5757--5823},
  year={2025},
  organization={PMLR}
}

@article{spigler2020asymptotic,
  title={Asymptotic learning curves of kernel methods: empirical data versus teacher--student paradigm},
  author={Spigler, Stefano and Geiger, Mario and Wyart, Matthieu},
  journal={Journal of Statistical Mechanics: Theory and Experiment},
  volume={2020},
  number={12},
  pages={124001},
  year={2020},
  publisher={IOP Publishing}
}

@article{gerbelot2023graph,
  title={Graph-based approximate message passing iterations},
  author={Gerbelot, C{\'e}dric and Berthier, Rapha{\"e}l},
  journal={Information and Inference: A Journal of the IMA},
  volume={12},
  number={4},
  pages={2562--2628},
  year={2023},
  publisher={Oxford University Press}
}

@article{loureiro2021learning,
  title={Learning curves of generic features maps for realistic datasets with a teacher-student model},
  author={Loureiro, Bruno and Gerbelot, Cedric and Cui, Hugo and Goldt, Sebastian and Krzakala, Florent and Mézard, Marc and Zdeborov{\'a}, Lenka},
  journal={Advances in Neural Information Processing Systems},
  volume={34},
  pages={18137--18151},
  year={2021}
}

@article{cheng2024dimension,
  title={Dimension free ridge regression},
  author={Cheng, Chen and Montanari, Andrea},
  journal={The Annals of Statistics},
  volume={52},
  number={6},
  pages={2879--2912},
  year={2024},
  publisher={Institute of Mathematical Statistics}
}

@article{cui2021generalization,
  title={Generalization error rates in kernel regression: The crossover from the noiseless to noisy regime},
  author={Cui, Hugo and Loureiro, Bruno and Krzakala, Florent and Zdeborov{\'a}, Lenka},
  journal={Advances in Neural Information Processing Systems},
  volume={34},
  pages={10131--10143},
  year={2021}
}

@article{erba2025nuclear,
  title={The nuclear route: Sharp asymptotics of erm in overparameterized quadratic networks},
  author={Erba, Vittorio and Troiani, Emanuele and Zdeborov{\'a}, Lenka and Krzakala, Florent},
  journal={Advances in Neural Information Processing Systems},
  volume={38},
  pages={88862--88901},
  year={2026}
}

@article{arous2025learning,
  title={Learning quadratic neural networks in high dimensions: SGD dynamics and scaling laws},
  author={Ben Arous, Gerard and Erdogdu, Murat and Vural, Nuri Mert and Wu, Denny},
  journal={Advances in Neural Information Processing Systems},
  volume={38},
  pages={146725--146812},
  year={2026}
}

@article{huang2018mesoscopic,
  title={Mesoscopic perturbations of large random matrices},
  author={Huang, Jiaoyang},
  journal={Random Matrices: Theory and Applications},
  volume={7},
  number={02},
  pages={1850004},
  year={2018},
  publisher={World Scientific}
}

@article{cagnetta2026deriving,
  title={Deriving Neural Scaling Laws from the statistics of natural language},
  author={Cagnetta, Francesco and Ravent{\'o}s, Allan and Ganguli, Surya and Wyart, Matthieu},
  journal={arXiv preprint arXiv:2602.07488},
  year={2026}
}

@article{yang2021tuning,
  title={Tuning large neural networks via zero-shot hyperparameter transfer},
  author={Yang, Ge and Hu, Edward and Babuschkin, Igor and Sidor, Szymon and Liu, Xiaodong and Farhi, David and Ryder, Nick and Pachocki, Jakub and Chen, Weizhu and Gao, Jianfeng},
  journal={Advances in Neural Information Processing Systems},
  volume={34},
  pages={17084--17097},
  year={2021}
}

@inproceedings{bordelon2023depthwise,
  title={Depthwise hyperparameter transfer in residual networks: Dynamics and scaling limit},
  author={Bordelon, Blake and Noci, Lorenzo and Li, Mufan and Hanin, Boris and Pehlevan, Cengiz},
  booktitle={12th International Conference on Learning Representations, ICLR 2024},
  year={2024}
}

@article{chizat2024feature,
  title={The feature speed formula: a flexible approach to scale hyper-parameters of deep neural networks},
  author={Chizat, L{\'e}na{\"\i}c and Netrapalli, Praneeth},
  journal={Advances in Neural Information Processing Systems},
  volume={37},
  pages={62362--62383},
  year={2024}
}

@article{chaintron2026resnets,
  title={Resnets of all shapes and sizes: Convergence of training dynamics in the large-scale limit},
  author={Chaintron, Louis-Pierre and Chizat, L{\'e}na{\"\i}c and Maas, Javier},
  journal={arXiv preprint arXiv:2603.18168},
  year={2026}
}

@article{defilippis2026scaling,
  title={Scaling laws and spectra of shallow neural networks in the feature learning regime},
  author={Defilippis, Leonardo and Xu, Yizhou and Girardin, Julius and Troiani, Emanuele and Erba, Vittorio and Zdeborov{\'a}, Lenka and Loureiro, Bruno and Krzakala, Florent},
  note= {ICLR},
  journal={arXiv preprint arXiv:2509.24882},
  year={2026}
}

@article{defilippis2026optimal,
  title={Optimal scaling laws in learning hierarchical multi-index models},
  author={Defilippis, Leonardo and Krzakala, Florent and Loureiro, Bruno and Maillard, Antoine},
  journal={arXiv preprint arXiv:2602.05846},
  year={2026}
}

@article{maillard2024bayes,
  title={Bayes-optimal learning of an extensive-width neural network from quadratically many samples},
  author={Maillard, Antoine and Troiani, Emanuele and Martin, Simon and Krzakala, Florent and Zdeborov{\'a}, Lenka},
  journal={Advances in Neural Information Processing Systems},
  volume={37},
  pages={82085--82132},
  year={2024}
}

@article{defilippis2024dimension,
  title={Dimension-free deterministic equivalents and scaling laws for random feature regression},
  author={Defilippis, Leonardo and Loureiro, Bruno and Misiakiewicz, Theodor},
  journal={Advances in Neural Information Processing Systems},
  volume={37},
  pages={104630--104693},
  year={2024}
}

@article{misiakiewicz2024non,
  title={A non-asymptotic theory of kernel ridge regression: deterministic equivalents, test error, and {GCV} estimator},
  author={Misiakiewicz, Theodor and Saeed, Basil},
  journal={arXiv preprint arXiv:2403.08938},
  year={2024}
}

@article{brown2020language,
  title={Language models are few-shot learners},
  author={Brown, Tom and Mann, Benjamin and Ryder, Nick and Subbiah, Melanie and Kaplan, Jared D and Dhariwal, Prafulla and Neelakantan, Arvind and Shyam, Pranav and Sastry, Girish and Askell, Amanda and others},
  journal={Advances in neural information processing systems},
  volume={33},
  pages={1877--1901},
  year={2020}
}

@article{kaplan2020scaling,
  title={Scaling laws for neural language models},
  author={Kaplan, Jared and McCandlish, Sam and Henighan, Tom and Brown, Tom B and Chess, Benjamin and Child, Rewon and Gray, Scott and Radford, Alec and Wu, Jeffrey and Amodei, Dario},
  journal={arXiv preprint arXiv:2001.08361},
  year={2020}
}

@article{hoffmann2022empirical,
  title={An empirical analysis of compute-optimal large language model training},
  author={Hoffmann, Jordan and Borgeaud, Sebastian and Mensch, Arthur and Buchatskaya, Elena and Cai, Trevor and Rutherford, Eliza and de Las Casas, Diego and Hendricks, Lisa Anne and Welbl, Johannes and Clark, Aidan and others},
  journal={Advances in neural information processing systems},
  volume={35},
  pages={30016--30030},
  year={2022}
}

@article{paquette2024four,
  title={4+3 phases of compute-optimal neural scaling laws},
  author={Paquette, Elliot and Paquette, Courtney and Xiao, Lechao and Pennington, Jeffrey},
  journal={Advances in Neural Information Processing Systems},
  volume={37},
  pages={16459--16537},
  year={2024}
}

@article{bahri2024explaining,
  title={Explaining neural scaling laws},
  author={Bahri, Yasaman and Dyer, Ethan and Kaplan, Jared and Lee, Jaehoon and Sharma, Utkarsh},
  journal={Proceedings of the National Academy of Sciences},
  volume={121},
  number={27},
  pages={e2311878121},
  year={2024},
  publisher={National Academy of Sciences}
}

@article{maloney2022solvable,
  title={A solvable model of neural scaling laws},
  author={Maloney, Alexander and Roberts, Daniel A and Sully, James},
  journal={arXiv preprint arXiv:2210.16859},
  year={2022}
}

@inproceedings{bordelon2020spectrum,
  title={Spectrum dependent learning curves in kernel regression and wide neural networks},
  author={Bordelon, Blake and Canatar, Abdulkadir and Pehlevan, Cengiz},
  booktitle={International Conference on Machine Learning},
  pages={1024--1034},
  year={2020},
  organization={PMLR}
}

@inproceedings{bordelon2024dynamical,
  title={A dynamical model of neural scaling laws},
  author={Bordelon, Blake and Atanasov, Alexander and Pehlevan, Cengiz},
  booktitle={Proceedings of the 41st International Conference on Machine Learning},
  pages={4345--4382},
  year={2024}
}

@article{atanasov2024scaling,
  title={Scaling and renormalization in high-dimensional regression},
  author={Atanasov, Alexander and Zavatone-Veth, Jacob A and Pehlevan, Cengiz},
  journal={Journal of Statistical Mechanics: Theory and Experiment},
  volume={2026},
  number={4},
  pages={043404},
  year={2026},
  publisher={IOP Publishing}
}

@article{caponnetto2007optimal,
  title={Optimal rates for the regularized least-squares algorithm},
  author={Caponnetto, Andrea and De Vito, Ernesto},
  journal={Foundations of Computational Mathematics},
  volume={7},
  number={3},
  pages={331--368},
  year={2007},
  publisher={Springer}
}

@inproceedings{martin2024impact,
  title={On the impact of overparameterization on the training of a shallow neural network in high dimensions},
  author={Martin, Simon and Bach, Francis and Biroli, Giulio},
  booktitle={International Conference on Artificial Intelligence and Statistics},
  pages={3655--3663},
  year={2024},
  organization={PMLR}
}

@article{simon2023eigenlearning,
  title={The eigenlearning framework: A conservation law perspective on kernel ridge regression and wide neural networks},
  author={Simon, James B and Dickens, Madeline and Karkada, Dhruva and DeWeese, Michael R},
  journal={Transactions on Machine Learning Research},
  year={2023}
}

@inproceedings{ren2025emergence,
  title={Emergence and scaling laws in SGD learning of shallow neural networks},
  author={Ren, Yunwei and Nichani, Eshaan and Wu, Denny and Lee, Jason D},
  booktitle={The Thirty-ninth Annual Conference on Neural Information Processing Systems},
  year={2025}
}

@book{mezard1988spin,
  title={Spin Glass Theory And Beyond: An Introduction To The Replica Method And Its Applications},
  author={Mézard, Marc and Parisi, Giorgio and Virasoro, Miguel Angel},
  volume={9},
  year={1987},
  publisher={World Scientific Publishing Company}
}

@inproceedings{boncoraglio2026singleheadattentionhighdimensions,
  title={Single-Head Attention in High Dimensions: A Theory of Generalization, Weights Spectra, and Scaling Laws},
  author={Boncoraglio, Fabrizio and Erba, Vittorio and Troiani, Emanuele and Xu, Yizhou and Krzakala, Florent and Zdeborov{\'a}, Lenka},
  booktitle={ICML},
  note = {arxiv:2509.24914},
  year={2026}
}

@article{martin2026high,
  title={High-dimensional analysis of gradient flow for extensive-width quadratic neural networks},
  author={Martin, Simon and Biroli, Giulio and Bach, Francis},
  journal={arXiv preprint arXiv:2601.10483},
  year={2026}
}

@article{Eckart_Young_1936, title={The Approximation of One Matrix by Another of Lower Rank}, volume={1}, DOI={10.1007/BF02288367}, number={3}, journal={Psychometrika}, author={Eckart, Carl and Young, Gale}, year={1936}, pages={211–218}}

@article{vilucchio2025asymptotics,
  title={Asymptotics of non-convex generalized linear models in high-dimensions: A proof of the replica formula},
  author={Vilucchio, Matteo and Dandi, Yatin and Rossignol, Mat{\'e}o Pirio and Gerbelot, Cedric and Krzakala, Florent},
  journal={arXiv preprint arXiv:2502.20003},
  year={2025}
}

@article{venturi19,
  title = {Spurious Valleys in One-Hidden-Layer Neural Network Optimization Landscapes},
  author = {Venturi, Luca and Bandeira, Afonso S. and Bruna, Joan},
  year = {2019},
  journal = {Journal of Machine Learning Research},
  volume = {20},
  number = {133},
  pages = {1--34}
}

@article{paszke2019pytorch,
  title={Pytorch: An imperative style, high-performance deep learning library},
  author={Paszke, Adam and Gross, Sam and Massa, Francisco and Lerer, Adam and Bradbury, James and Chanan, Gregory and Killeen, Trevor and Lin, Zeming and Gimelshein, Natalia and Antiga, Luca and others},
  journal={Advances in neural information processing systems},
  volume={32},
  year={2019}
}

@InProceedings{pmlr-v80-du18a,
  title = 	 {On the Power of Over-parametrization in Neural Networks with Quadratic Activation},
  author =       {Du, Simon and Lee, Jason},
  booktitle = 	 {Proceedings of the 35th International Conference on Machine Learning},
  pages = 	 {1329--1338},
  year = 	 {2018},
  editor = 	 {Dy, Jennifer and Krause, Andreas},
  volume = 	 {80},
  series = 	 {Proceedings of Machine Learning Research},
  month = 	 {10--15 Jul},
  publisher =    {PMLR},
}

@article{soltanolkotabi2018theoretical,
  title={Theoretical insights into the optimization landscape of over-parameterized shallow neural networks},
  author={Soltanolkotabi, Mahdi and Javanmard, Adel and Lee, Jason D},
  journal={IEEE Transactions on Information Theory},
  volume={65},
  number={2},
  pages={742--769},
  year={2018},
  publisher={IEEE}
}

@article{zdeborova2016statistical,
  title={Statistical physics of inference: Thresholds and algorithms},
  author={Zdeborov{\'a}, Lenka and Krzakala, Florent},
  journal={Advances in Physics},
  volume={65},
  number={5},
  pages={453--552},
  year={2016},
  publisher={Taylor \& Francis}
}

@article{advani2013statistical,
  title={Statistical mechanics of complex neural systems and high dimensional data},
  author={Advani, Madhu and Lahiri, Subhaneil and Ganguli, Surya},
  journal={Journal of Statistical Mechanics: Theory and Experiment},
  volume={2013},
  number={03},
  pages={P03014},
  year={2013},
  publisher={IOP Publishing and SISSA}
}

@book{mezard2009information,
  title={Information, physics, and computation},
  author={Mézard, Marc and Montanari, Andrea},
  year={2009},
  publisher={Oxford University Press}
}

@article{gamarnikStationaryPointsShallow2020,
  title={Stationary points of shallow neural networks with quadratic activation function},
  author={Gamarnik, David and K{\i}z{\i}lda{\u{g}}, Eren C and Zadik, Ilias},
  journal={arXiv preprint arXiv:1912.01599},
  year={2019}
}

@article{bayati2010lasso,
  title={The LASSO risk: asymptotic results and real world examples},
  author={Bayati, Mohsen and Pereira, Jos{\'e} and Montanari, Andrea},
  journal={Advances in Neural Information Processing Systems},
  volume={23},
  year={2010}
}

@article{bayati2011lasso,
  title={The LASSO risk for Gaussian matrices},
  author={Bayati, Mohsen and Montanari, Andrea},
  journal={IEEE Transactions on Information Theory},
  volume={58},
  number={4},
  pages={1997--2017},
  year={2011},
  publisher={IEEE}
}

@article{rangan2016fixed,
  title={Fixed points of generalized approximate message passing with arbitrary matrices},
  author={Rangan, Sundeep and Schniter, Philip and Riegler, Erwin and Fletcher, Alyson K and Cevher, Volkan},
  journal={IEEE Transactions on Information Theory},
  volume={62},
  number={12},
  pages={7464--7474},
  year={2016},
  publisher={IEEE}
}

@article{berthierStateEvolutionApproximate2020,
  title = {State Evolution for Approximate Message Passing with Non-Separable Functions},
  author = {Berthier, Rapha{\"e}l and Montanari, Andrea and Nguyen, Phan-Minh},
  year = {2020},
  month = mar,
  journal = {Information and Inference: A Journal of the IMA},
  volume = {9},
  number = {1},
  pages = {33--79},
  issn = {2049-8764, 2049-8772},
  copyright = {https://academic.oup.com/journals/pages/open\_access/funder\_policies/chorus/standard\_publication\_model},
  langid = {english}
}

@article{mirsky1975trace,
  title={A trace inequality of John von Neumann},
  author={Mirsky, Leon},
  journal={Monatshefte f{\"u}r mathematik},
  volume={79},
  number={4},
  pages={303--306},
  year={1975},
  publisher={Springer}
}

@article{montanari2012graphical,
  title={Graphical models concepts in compressed sensing.},
  author={Montanari, Andrea and Eldar, YC and Kutyniok, G},
  journal={Compressed Sensing},
  pages={394--438},
  year={2012}
}

@article{donoho2016high,
  title={High dimensional robust m-estimation: Asymptotic variance via approximate message passing},
  author={Donoho, David and Montanari, Andrea},
  journal={Probability Theory and Related Fields},
  volume={166},
  pages={935--969},
  year={2016},
  publisher={Springer}
}

@article{donoho13,
	title        = {The phase transition of matrix recovery from Gaussian measurements matches the minimax MSE of matrix denoising},
	author       = {Donoho, David L. and Gavish, Matan and Montanari, Andrea},
	year         = 2013,
	month        = may,
	journal      = {Proceedings of the National Academy of Sciences},
	publisher    = {Proceedings of the National Academy of Sciences},
	volume       = 110,
	number       = 21,
	pages        = {8405–8410},
	issn         = {1091-6490},
}

@inproceedings{fazel2008compressed,
  title={Compressed sensing and robust recovery of low rank matrices},
  author={Fazel, Maryam and Candes, Emmanuel and Recht, Ben and Parrilo, Pablo},
  booktitle={2008 42nd Asilomar Conference on Signals, Systems and Computers},
  pages={1043--1047},
  year={2008},
  organization={IEEE}
}

@article{arjevani2025geometry,
  title={Geometry and optimization of shallow polynomial networks},
  author={Arjevani, Yossi and Bruna, Joan and Kileel, Joe and Polak, Elzbieta and Trager, Matthew},
  journal={SIAM Journal on Applied Algebra and Geometry},
  volume={10},
  number={2},
  pages={174--209},
  year={2026},
  publisher={SIAM}
}

@article{baik2005phase,
  title={Phase transition of the largest eigenvalue for nonnull complex sample covariance matrices},
  author={Baik, Jinho and Ben Arous, G{\'e}rard and Peche, Sandrine},
  journal={Annals of probability},
  volume={33},
  number={5},
  pages={1643--1697},
  year={2005}
}
\appendix

\section{Derivation of Result \ref{res:se}: analytical test error characterization}
\label{app:derivation_SE}

We derive the characterization in Result \ref{res:se} by adapting the derivation of \cite{erba2025nuclear} to power-law targets and width-constrained students. 
For the sake of the derivation, as discussed in the main text, we first assume that $n=\Theta(d^2)$, $p=\Theta(d)$, $\tl=\Theta(1)$, that $\bS_\star$ has a limiting spectral distribution for $d \gg 1$. We proceed under the \emph{Replica Symmetric} (RS) assumption, a standard hypothesis in the statistical physics literature \cite{mezard1988spin,mezard2009information}. In the present setting, this assumption is equivalent to postulating that a rank-constrained extension of the Approximate Message Passing (AMP) algorithm introduced in \cite{erba2025nuclear} tracks the global minimizer of the empirical risk minimization problem in the high-dimensional limit. This connection can be made rigorous when the loss is convex, as in the full-rank setting studied in \cite{erba2025nuclear}. In our rank-constrained formulation, however, the ERM problem is generally non-convex, and validating the replica-symmetric description requires additional checks, which we discuss below.
In a second stage we will then assume that the derived equations can be relaxed to general scaling, and verify this by comparing with numerical experiments.

\paragraph{Gaussian universality and AMP.}
The derivation proceeds in two steps:
\begin{itemize}
    \item 
    {\bf Gaussian universality:} one shows that for the asymptotic behavior of the global minimizer, the training data $\bG_\mu = (\bx_\mu \bx_\mu^\top - \bI_d)/\sqrt{d}$ can be replaced with surrogate Gaussian data, i.e. $\boldsymbol{\tilde{G}}_\mu \sim \text{GOE}(d)$. With this substitution, \eqref{eq:def:erm_equiv_intro} becomes a Gaussian linear regression problem with matrix weights and data, plus a non-separable $\ell_1$+PSD+rank-constraint regularization.
    \citep{maillard2024bayes,xu2025fundamental}
    \item 
    {\bf AMP and its analysis:}
    The asymptotic behavior of ERM in Gaussian linear models has been studied extensively  \citep{bayati2010lasso,montanari2012graphical,rangan2016fixed,loureiro2021learning,vilucchio2025asymptotics}, in particular through Approximate Message Passing (AMP). This framework allows to write an algorithm, the mentioned AMP, whose fixed points are stationary points of the original ERM loss, and that can be studied analytically for $d\gg 1$ through the state evolution framework. Crucially, AMP and state evolution can deal with non-separable regularization \citep{berthierStateEvolutionApproximate2020,gerbelot2023graph}, such as the one we need to treat here. 
\end{itemize}

We do not repeat here the full derivation, as it is quite standard. We just mention where the role of the rank-constraint enters it, and what changes does it induce.
For an ERM problem of the form \eqref{eq:def:erm_equiv_intro}
with the data model \eqref{eq:def:target}, 
\cite{erba2025nuclear} shows that the asymptotic behavior of the global minimum can be determined by solving the system of equations
\begin{equation}
\label{eq:amp-se}
\begin{aligned}
\begin{cases}
\displaystyle
\widehat{\Sigma}
= \frac{2n}{d^2}\,\frac{1}{\Sigma + \frac14},
\\[0.75em]
\displaystyle
\widehat{m}
= \frac{2n}{d^2}\,\frac{1}{\Sigma + \frac14},
\\[0.75em]
\displaystyle
\widehat{q}
= \frac{2n}{d^2}\,
\frac{Q_\star - 2m + q + \frac{\Delta}{2}}
{\left(\Sigma + \frac14\right)^2},
\end{cases}
\qquad
\begin{cases}
\displaystyle
m
= -2\,\partial_{\widehat{m}}
\Psi(\widehat{\Sigma},\widehat{q},\widehat{m}),
\\[0.75em]
\displaystyle
q
= 4\,\partial_{\widehat{\Sigma}}
\Psi(\widehat{\Sigma},\widehat{q},\widehat{m}),
\\[0.75em]
\displaystyle
\Sigma
= -4\,\partial_{\widehat{q}}
\Psi(\widehat{\Sigma},\widehat{q},\widehat{m}) .
\end{cases}
\end{aligned}
\end{equation}
where
\begin{equation}
\Psi(\hSigma, \hq, \hm) = 
    \frac{1}{d} \mathbb{E}_{\bZ}
        \min_{\bS \succeq 0 , \, \text{rk}(\bS) \leq p}
        \left[ 
        \tl \Tr(\bS) + \frac{\hSigma}{4} \| \bS \|_F^2 - \frac{1}{2} \Tr( \bS^\top ( \hm \bS_\star + \sqrt{\hq} \bZ ) )
        \right] \, ,
\end{equation} 
and $\bZ \sim \text{GOE}(d)$.
This requires the computation of the minimizer (which we will call also denoiser)
\begin{equation}\label{eq:denoiser}
     \bD(\bY) = \underset{\bS \succeq 0 , \, \text{rk}(\bS) \leq p}{\text{argmin}}
        \left[ 
        \tl  \Tr(\bS) + \frac{\hSigma}{4} \| \bS \|_F^2 - \frac{1}{2} \Tr( \bS^\top \bY )
        \right] 
\end{equation}
for a fixed symmetric matrix $\bY \in \bbR^{d \times d}$. 
Notice that $m$ and $q$ have a clear interpretation. For any fixed value of $\hq, \hm, \hSigma, p$ and $\bP = \bD(\hm \bS_\star + \sqrt{\hq} \bZ)$, we have that \cite[Appendix A.4.3]{erba2025nuclear}
\begin{equation}
    \begin{split}
            m&
            = \frac{1}{d}{\mathbb E}_{\bZ} \left[\Tr(\bP^\top\bS_\star )\right]
            = -2\,\partial_{\widehat{m}} \Psi(\widehat{\Sigma},\widehat{q},\widehat{m}),
            \\
            q&= \frac{1}{d}{\mathbb E}_{\bZ} \left[
             \Tr(\bP^\top\bP )\right]
             = 4\,\partial_{\widehat{\Sigma}} \Psi(\widehat{\Sigma},\widehat{q},\widehat{m}),
             \\
    \end{split}
\end{equation}
allowing to interpret $m$ as an overlap between a ground truth matrix $\bS_\star$ and a noisy version of it $\hm \bS_\star + \sqrt{\hq} \bZ$ denoised through $\bD(\cdot)$, and $q$ as its Frobenius norm squared.
Consequently, we can also write the test error for any scalar multiple of such matrix as
\begin{equation}\label{app:test_generic}
    \begin{split}
        \caE(c \bP) 
    &= \frac{1}{d} \| c \bP - \bS_\star \|_F^2 
    = Q_\star - 2 c m + q c^2
    = Q_\star 
    + 4 c \,\partial_{\widehat{m}} \Psi
    + 4 c^2 \,\partial_{\widehat{\Sigma}} \Psi
    \end{split}
\end{equation}
where $Q_\star = \Tr(\bS^2_\star) / d$ and $c \in \bbR$

\textbf{Computation of the denoiser and role of rank constraint.}
We now compute \eqref{eq:denoiser}.
We first remark that any $d\times d$ PSD matrix with rank at most $p < d$ can be decomposed spectrally as $\bS = \bO \bL \bO^\top$, where $\bO$ is a $d\times d$ rotation matrix and  $\bL = \text{diag}(L_1, \dots, L_p, 0, \dots, 0)$ with $L_1 \geq L_2 \geq \dots \geq L_p \geq 0$.
Thus
\begin{equation}
     \bD(\bY) = \underset{L_1 \geq \dots \geq L_p \geq 0}{\text{argmin}}
        \left[ 
        \tl \sum_{i=1}^p L_i + \frac{\hSigma}{4} \sum_{i=1}^p L_i^2 
        - \frac{1}{2} \underset{\bO \text{ rotation}}{\text{argmax}}
        \Tr( \bO^\top \bL \bO \bY )
        \right] \, .
\end{equation}
By Von Neumann's inequality \citep{mirsky1975trace}, 
\begin{equation}
    \underset{\bO \text{ rotation}}{\text{argmax}}
        \Tr( \bO^\top \bL \bO \bY )
        = \sum_{i=1}^p L_i Y_i
\end{equation}
where $Y_i$ are the eigenvalues of $\bY$ sorted decreasingly, giving
\begin{equation}
     \bD(\bY) = \underset{L_1 \geq \dots \geq L_p \geq 0}{\text{argmin}}
        \left[ 
        \tl \sum_{i=1}^p L_i + \frac{\hSigma}{4} \sum_{i=1}^p L_i^2 
        - \frac{1}{2} \sum_{i=1}^p L_i Y_i
        \right] \, .
\end{equation}
The minimization can be now solved independently for all $i=1, \dots, p$ over the set $L_i \geq 0$, as the objective is a sum over functions depending on a single value of $i$, and checking a posteriori that the overall monotonicity constraint $L_1 \geq \dots \geq L_p$ is satisfied. The fixed $i$ solution gives
\begin{equation}
    L_i = \frac{1}{\hSigma} \text{ReLU} (Y_i - 2\tl)
\end{equation}
from which we see that all the $L_i$ are automatically sorted decreasing due to the sorting of the $Y_i$.
Plugging back in we obtain
\begin{equation}\label{eq:finalapp}
\begin{split}
    \Psi(\hSigma, \hq, \hm) &= 
        - \frac{\hm^2}{4\hSigma}
        \mathbb{E}_{\bZ \sim \text{GOE}(d)}
        \frac{1}{d} \sum_{i=1}^p \text{ReLU} (\nu_i - 2\tl/\hm) ^2
        =  - \frac{\hm^2}{4\hSigma} J_p\left( \frac{\sqrt{\hq}}{\hm}, \frac{2\tl}{\hm}\right)
     \, ,
\end{split}
\end{equation} 
where we called $\nu_i$ the $i$-th eigenvalue sorted decreasingly of $\bS_\star + \sqrt{\hq}/\hm \bZ$, and finally used the notation of Result \ref{res:se}.
Algebraic manipulations of \eqref{eq:amp-se} under the change of variable $\delta = \sqrt{\hq}/\hm$ and $\epsilon = 2 / \hm$ leads to \eqref{eq:SE_ERM}.
We thus see that the rank constraint on the original problem induces an additional spectral cut in \eqref{eq:finalapp} (the sum going up to $p < d$, and not up to $d$ as it would in the non-rank-constrained case). 

This concludes the raw derivation of \eqref{eq:SE_ERM}. 
The expression of the training loss, test error, and spectral distribution follow from the AMP/state evolution derivation \cite{erba2025nuclear}, and are given by
\begin{equation}
    \text{sp} \left[\bhS\right] \overset{d}{\sim} \text{sp}\left[ (\bS_\star + \delta \bZ - \tl \epsilon \bI_d)^+_{(p)} \right]
    \, , \quad
    \caE(\bhW) \sim \frac{2n}{d^2}\delta^2 - \frac{\Delta}{2} 
    \, , \quad
    \caL(\bhW) \sim
    \frac{\delta^2}{4\epsilon^2} + \frac{\tl}{d} \Tr(\bhS)
    \, ,
\end{equation}
where by $\text{sp}[ \cdot ]$ we mean the empirical spectral distribution. Moreover, combining \eqref{app:test_generic} and \eqref{eq:finalapp} with $\bP = (\hm / \hSigma) \bhS$ and $(\bS_\star + \delta \bZ - \tl \epsilon \bI_d)^+_{(p)}$, we get the alternative expression
\begin{equation}
    \label{eq:test-different}
    \caE(\bhW) \sim 
    Q_\star + (\delta \del_1 + \tl \epsilon \del_2 - 1) J_p (\delta, \tl \epsilon) \, .
\end{equation}

\textbf{Validity.}
Now, as discussed in the main text, we take two leaps of faith. 
First, we assume that this derivation holds for different scaling regimes than the one in which these equations are usually derived or proven.
Second, we assume that the non-convexity of the loss does not break the replica symmetric assumption 
The first point is tricky to control analytically, and we are not aware of any line of attack. 
The second point can be approached (in the strict asymptotic scaling for which the derivation formally holds) by checking the necessary (but not sufficient)
replicon condition (see \cite{vilucchio2025asymptotics} for a related discussion)
\begin{equation} 
  \frac{1}{d^2}
    \left[2 \sum_{i=1}^d (\del \eta_i(\nu_i))^2
        +
        \sum_{i \neq i'} \left(\frac{\eta_i(\nu_i) - \eta_{i'}(\nu_{i'})}{\nu_i - \nu_{i'}} \right)^2
    \right]<\frac{2n}{d^2} \, ,
\label{eq:RS}
\end{equation}
where
\begin{equation}
\eta_i(x)= \theta( 1 \leq i \leq p) \text{ReLU} (\nu_i - 2\tl/\hm) \, .
\end{equation}
This condition is related to the linear stability of AMP at its fixed point. In order to state a sufficient condition for the replica symmetric assumption, we would have to work out the so-called replica symmetry breaking version of the analysis, as known in the physics of disordered systems \cite{mezard1988spin,mezard2009information}.
In Appendix \ref{app:derivation_scaling}, we discuss the replicon condition for each phase we consider in the main text, but we stress here that as soon as we enter the non-asymptotic scaling domain, there is no reason to believe that this condition would still be meaningful in the usual sense.

\textbf{Pruning.}
The test error of the post-training pruning procedure sketched in the main text can be studied by setting $p=d$ (training at full width) 
and projecting the resulting global minimizer to matrices with rank lower than $p_{\rm pruning} < d$. Thus, we want to compute the generalization of $\bhS_{\rm pruning} = (\bS_\star + \delta \bZ - \tl \epsilon \bI_d)^+_{p_{\rm pruning}}$, where $\delta, \epsilon$ solve \eqref{eq:SE_ERM} for $p=d$. This can be computed using \eqref{app:test_generic} and \eqref{eq:finalapp} with $\bP = (\hm / \hSigma) \bhS_{\rm pruning}$, giving
\begin{equation}\label{app:test-prune}
    \caE(\bhW) \sim 
    Q_\star 
    + 4 (\hm / \hSigma)^{-1} \,\partial_{\widehat{m}} \Psi
    + 4 (\hm / \hSigma)^{-2} \,\partial_{\widehat{\Sigma}} \Psi
    =
    Q_\star + (\delta \del_1 + \tl \epsilon \del_2 - 1) J_{p_{\rm pruning}} (\delta, \tl \epsilon)
\end{equation}
This is equivalent to substituting $\tl \epsilon$ at the solution of \eqref{eq:SE_ERM} with $\min(\tl \epsilon, x_0)$ where $x_0$ is the value of the $p$-th eigenvalue of $\hS$, sorted decreasingly.

\section{Derivation of the scaling laws}

Results \ref{res:under-reg}, \ref{res:over-reg} are derived from Result \ref{res:se} by careful asymptotic analysis of \eqref{eq:SE_ERM}. For notational simplicity we assume that $\bS_{\star}$ 
has eigenvalues $\{\sqrt{d}\, i^{-\gamma} \}_{i=1}^d$ with $\gamma > 1/2$, which can me mapped to the minimization objective in the main text \eqref{eq:def:erm_equiv_intro} by rescaling $\Delta Q_\star \to \Delta $ and $\lambda \to \lambda / Q_\star$, where $Q_\star:=\zeta(2\gamma)$ alters only the asymptotic constants, not the scaling. The learned matrix $\bhS$ in the appendix will be a factor $Q_\star$ larger than in the main, as it will be the case for the generalization error and the loss.

The scaling laws obtained in this Appendix are a result of a simplifying assumption on the spectrum of the learned features $\bhS$, which we do coherently with \cite{defilippis2026scaling}: calling $\mu_\delta(\cdot)$ the spectral density of $\bS_\star + \delta \bZ $, we approximate it as a semicircular bulk of radius $2\delta$ of relative mass $d - K $ plus $K$ outlying spikes.
More explicitly, we model the eigenvalues of $\bS_\star + \delta \bZ $ as $\{\xi_i\}_{i=1,...d-K}\cup\{\nu_i\}_{i=1,...,K}$, where we assume that $\xi_{i}\overset{\mathrm{i.i.d.}}{\sim} \mu_{\rm sc}(x/\delta)/\delta$, with $\mu_{\rm sc}(\cdot)$ the standard semi-circle law of GOE matrices
\begin{equation}
    \mu_{\rm sc}(x) = \frac{\sqrt{4 - x^2}}{2\pi} \theta\big(|x|<2\big)\,,
\end{equation}
and the top eigenvalues are the pushforward of 
$\{\sqrt{d}\, i^{-\gamma} \}_{i=1}^K$
through the BBP mapping \cite{baik2005phase,huang2018mesoscopic} $f_\delta(x) = x + \delta^2 / x$:
\begin{equation}
    \nu_i = f_\delta\big(\sqrt{d}\, i^{-\gamma} \big)\,.
\end{equation}
The number of spikes outside the semicircle is fixed by imposing that $K$ is the largest integer such that 
\begin{equation}\label{eq:K}
    \sqrt{d}K^{-\gamma}\geq\delta
\end{equation}
following again BBP theory. We expect this approximation to be good as long as $K\ll d$.
In the following we will use the notation 
$\nu_{K+1} = \xi_{1} = 2\delta$
to indicate the largest eigenvalue in the semicircle. 
Based on this simplification, we write the spectral density $\mu_\delta(\cdot)$ as
\begin{equation}
\mu_\delta(x) \approx \left(1 - \frac{K}{d} \right)\mu_{\rm sc}(x/\delta) / \delta
    + \frac{K}{d} \sum_{i = 1}^{K} \delta\left( x - f_\delta\big(\sqrt{d}i^{-\gamma}\big)\right),
\end{equation}
where $\delta(\cdot)$ is the Dirac delta function. The spectrum of $\bhS = (\bS_\star + \delta \bZ - \tl \epsilon \bI_d)^+_{(p)} $ is thus given by
\begin{equation}
\mu(x)
=
\begin{cases}
\mu_\delta(x+\tl\epsilon), & x>x_0^+,\\
0, & x\le x_0^+,
\end{cases}
\qquad
x_0^+ := \max\{x_0,0\},
\end{equation}
in addition to a Dirac delta in zero such that $\int_\mathbb{R}\mu(x)\dd x=1$. $x_0$ is a cutoff imposing that the rank of $\bhS$ is at most $p$. In formulas this means $x_0$ is defined by imposing that $\bhS$ has at least $\frac{d-p}{d}$ zero eigenvalues, implying
\begin{equation}
\label{eq:cutoff}
x_0 = \inf \left[ t \, \Big| \,  \int_{t}^{\infty} \mu_\delta(x+\tl\epsilon)\,{\rm d} x
\leq \frac{p}{d} \right]
\end{equation}
As long as $x_0 \leq 0$, the rank constraint is not active as the PSD constraint already produces a matrix with low enough rank.
With this form of the spectrum in mind, we can then study \eqref{eq:SE_ERM} in various regimes. In Appendix \ref{app:derivation_scaling} we study the scaling regimes for all noisy data models $\Delta > 0$, leaving the case $\Delta = 0$ to Appendix \ref{app:noiseless}. In Appendix \ref{app:pruning} we describe the scaling regimes of pruning.

\paragraph{Replicon.} We discuss how the condition \eqref{eq:RS} can be specified for this target. Rewriting it in terms of density, it reads
\begin{equation}
\int dx\mu_\delta(x)\int dy\mu_\delta(y)\left(\frac{\eta(x)-\eta(y)}{x-y}\right)^2<\frac{2n}{d^2},
\label{eq:RS_continuous}
\end{equation}
where
\begin{equation}
\eta(x)
=
\begin{cases}
x-\tl\epsilon, & x-\tl\epsilon > x_0^+,\\
0, & x-\tl\epsilon \le x_0^+,
\end{cases}
\qquad
x_0^+ := \max\{x_0,0\}.
\end{equation}
We then have three cases for the behaviour of this condition:
\begin{itemize}
    \item When the rank-$p$ cutoff has no effect, i.e. $x_0<0$, the spectrum is the same as the full rank case \cite{erba2025nuclear,defilippis2026scaling}, and the replicon condition is always satisfied by the convexity of the full rank optimization problem.
    \item When the cutoff is inside the bulk, i.e., $0<\tl\epsilon+x_0^+\leq\delta$, the left side of \eqref{eq:RS} diverges as $\int_{1/d}^L r^{-1}\,{\rm d}r\sim\log d$ because the mean spacing between eigenvalues is approximately $d^{-1}$. For this case, the replicon condition is satisfied only if $n\gg d^2\log d$.

    \item When the cutoff is outside the bulk, i.e.
    $\nu_k<\tl\epsilon+x_0^+\leq\nu_{k+1}$ for some $0\leq k\leq K$, then the left side of \eqref{eq:RS} is of order
    \begin{equation}
    \frac{1}{d^2}\sum_{i\leq k<j}\left(\frac{\eta(\nu_i)-\eta(\nu_j)}{\nu_i-\nu_j}\right)^2\approx \frac{1}{d^2}\sum_{i\leq k<j}\left(\frac{\sqrt{d}i^{-\gamma}}{\sqrt{d}\gamma (i-j)i^{-\gamma-1}}\right)^2\approx\frac{k^2}{d^2\gamma}\log d.
    \end{equation}
    Then replicon condition \eqref{eq:RS} thus holds as long as $k^2\ll n/\log d$. This will be always true for the case $\gamma > 1/2$ we are analyzing, because by \cite[Theorem 3.2]{defilippis2026optimal} the number of spikes is upper bounded by $K\leq (n/d)^{1/2\gamma}$, and thus for $\gamma>1/2$ the replicon condition is always satisfied once the cutoff is outside the bulk.
\end{itemize}

\subsection{Derivation of the excess test error for width-constrained networks}
\label{app:derivation_scaling}
In this section we derive the phase diagram. We will always assume $\rho>0\implies p\gg1$.

We now proceed with the solution of the equations. As we derived in Appendix \ref{app:derivation_SE}, the excess test error is given by \eqref{eq:test-different}
\begin{equation}
    \caE(\bhW) \sim 
    Q_\star + (\delta \del_1 + \tl \epsilon \del_2 - 1) J_p (\delta, \tl \epsilon)\,,
\end{equation}
where $J_p (\delta, \tl \epsilon)$ is the second spectral moment of $\bhS$
\begin{equation}
     J_p(\delta,\tl\epsilon) = \frac{1}{d} \Tr\left( \bhS_{\delta,\tl\epsilon}^2 \right)\,.
\end{equation}
The solution will consist in enumerating all the possible configurations for the spectrum of $\bhS$ in terms of the magnitude of the spikes, the width of the bulk $\delta$ and the shift $\tl\epsilon$, and computing the leading behavior of $J_p$ in each regime. These conditions will be translated into conditions on the network width, dimension, regularization and the number of samples, and one can verify that we spanned the whole space of possibilities when considering the union of all these regions.\\

We start by remarking that if $x_0^+ = 0$, that is when the ERM solution already has a rank less than $p$, we will obtain the same scaling results from \citep{defilippis2026scaling}, since the limiting factor is not the width of the network but the regularization shift of the eigenvalues. We will thus focus on the cases $x_0^+ > 0$, in which the cutoff is active. \\

\noindent In the derivation, we will make use of the approximations valid for $s\ll 1$
\begin{equation} \label{eq:bulk_int_approx}
    \int_{2-s}^2 \mu_{\rm sc}(x)\,{\rm d}x= \frac{2}{3\pi} s^{3/2}-\frac{1}{20\pi} s^{5/2}+\caO(s^{7/2})\,,
\end{equation}
as well as
 \begin{equation}\label{eq:bulk_J_approx}
 \begin{split}
\int_{2-s}^2 (x-a)^2\,\mu_{\rm sc}(x)\,{\rm d}x 
&= \frac{2}{3\pi}(2-a)^2 s^{3/2} \quad + \left[-\frac{4}{5\pi}(2-a) - \frac{(2-a)^2}{20\pi}\right]s^{5/2} \\
&\quad + \left[\frac{2}{7\pi} + \frac{(2-a)}{14\pi} - \frac{(2-a)^2}{448\pi}\right]s^{7/2}+  \mathcal{O}(s^{9/2})
 \end{split}
\end{equation}
We define the two natural parts into which $J_p$ can be decomposed: one part for the semicircle contribution ($J_p^{(1)}$), and one for the spike contribution ($J_p^{(2)}$)
\begin{align}
    \begin{dcases}
        J_p^{(1)}(\delta,\tl\epsilon)=\int_{\tl\epsilon+x_0^+}^{2\delta}(x-\tl\epsilon)^2\mu_{\rm sc,\delta}(x)\dd x=\delta^2\int_{2-s}^{2}\left(x-\frac{\tl\epsilon}{\delta}\right)^2\mu_{\rm sc}(x)\dd x\\
        J_p^{(2)}(\delta,\tl\epsilon)=\frac{1}{d}\sum_{i=1}^{\min(K,p)}\left[\sqrt{d}i^{-\gamma}+\frac{\delta^2}{\sqrt{d}i^{-\gamma}}-\tl\epsilon\right]^2
    \end{dcases}
\end{align}
where $s:=2-\frac{\tl\epsilon+x_0^+}{\delta}$ and $1 \leq K \leq d$ is the largest integer such that $\sqrt{d}K^{-\gamma}\geq\delta$. 

\paragraph{Non-rank-constrained phases.}
Before proceeding, we first recall the phase diagram from \cite{defilippis2026scaling} (with $\lambda$ replaced by $\tl$). Once the $p$ is larger than the rank of the non-constrained ERM solution, which we compute for each phase below, the rank constraint is inactive and we obtain the same error rates.
In general, the rank of the ERM can be computed by summing the amount of spikes outside of the bulk of the spectrum, plus the bulk contribution.
\begin{itemize}
    \item \textbf{Phase Ia}: the excess error is given by
    \begin{equation}
    \caE=Q^\star \mathif \tl\ll\sqrt{\frac{n}{d^2}}\mathand n\ll d\mathand p\geq d.
    \end{equation}
    The rank of the ERM solution is given by
    \begin{equation}
    d\int_{\tl\epsilon}^{2\delta}\mu_{\rm sc,\delta}(x)dx\approx\frac{2d}{3\pi}\left(2-\frac{\tl\epsilon}{\delta}\right)^{3/2}\approx\frac{2d}{3\pi}\left(\frac{15\pi n}{4d^2}\right)^{3/5}=\Theta(n^{3/5}d^{-1/5}),
    \label{eq:rank-ERM}
    \end{equation}
    where we use $2-\frac{\tl\epsilon}{\delta}\approx\left(\frac{15\pi n}{4d^2}\right)^{2/5}\ll1$ from \cite{defilippis2026scaling} and \eqref{eq:bulk_int_approx}. Therefore, we can obtain
    \begin{equation}
    \boxed{
    \caE=Q^\star \mathif \tl\ll\sqrt{\frac{n}{d^2}}\mathand n\ll d\mathand p\gg n^{3/5}d^{-1/5}}.
    \label{eq:phaseIa}
    \end{equation}
    This corresponds to the yellow region on the left side of the dashed line in Figure \ref{fig:fig2}(a).

    \item \textbf{Phase Ib}: the excess error is given by
    \begin{equation}
    \caE=Q^\star \mathif \tl\gg\sqrt{\frac{n}{d^2}}\mathand n\gg\tl d^{3/2}\mathand p\geq d.
    \end{equation}
    In this phase the rank of the ERM solution is zero, the rank constraint is always inactive, and we can write
    \begin{equation}
    \boxed{\caE=Q^\star \mathif \tl\gg\sqrt{\frac{n}{d^2}}\mathand n\gg\tl d^{3/2}},
    \end{equation}
    or equivalently
    \begin{equation}
    \boxed{\beta=0\mathif\alpha<\ell+\frac{3}{2}\mathand\ell>\frac{\alpha}{2}-1}.
    \end{equation}
    This corresponds to the yellow region in Figure \ref{fig:fig2}(b).

    \item \textbf{Phase II}: the excess error is given by
    \begin{equation}
    \caE=\frac{2\gamma}{2\gamma-1}\left(\tl\frac{d^{3/2}}{4n}\right)^{\frac{2\gamma-1}{\gamma}}\mathif\max\left(\sqrt{\frac{n}{d^2}},\frac{n}{d^{3/2+\gamma}}\right)\ll\tl\ll\frac{n}{d^{3/2}}\mathand p\geq d.
    \end{equation}
    Moreover, there are $K=(4n/\tl d^{3/2})^{1/\gamma}$ spikes, and thus if $p\gg K\implies \tl\gg\frac{n}{d^{3/2}p^\gamma}$ the results remain unchanged. Therefore, we have
    \begin{equation}
    \boxed{\caE=\frac{2\gamma}{2\gamma-1}\left(\tl\frac{d^{3/2}}{4n}\right)^{\frac{2\gamma-1}{\gamma}}\mathif\max\left(\sqrt{\frac{n}{d^2}},\frac{n}{d^{3/2}p^\gamma},\frac{n}{d^{3/2+\gamma}}\right)\ll\tl\ll\frac{n}{d^{3/2}}}.
    \end{equation}
    Equivalently, we can write
    \begin{equation}
    \beta=-\left(\alpha-\ell-\frac{3}{2}\right)\left(2-\frac{1}{\gamma}\right)\mathif\ell+\frac{3}{2}<\alpha<\ell+\frac{3}{2}+\gamma\max(\rho,1).
    \end{equation}
    This corresponds to the blue region in Figure \ref{fig:fig2}(b).

    \item \textbf{Phase III}:  all spikes are recovered, thus the ERM solution has full rank, and the excess error is 
    \begin{equation}
    \boxed{\caE=\frac{\tl^2d^4}{16n^2}\mathif\tl\gg\sqrt{\frac{n}{d^2}}\mathand n\gg\tl d^{3/2+\gamma}\mathand p\geq d},
    \end{equation}
    or equivalently
    \begin{equation}
    \boxed{\beta=-2(\alpha-\ell-2)\mathif\ell>\frac{\alpha}{2}-1\mathand\alpha>\ell+\frac{3}{2}+\gamma\mathand\rho\geq1}.
    \end{equation}
    This corresponds to the green region in Figure \ref{fig:fig2}(b).

    \item \textbf{Phase IV and V}: the excess error is given by
    \begin{equation}
    \begin{aligned}\caE=\frac{24\gamma^3}{4\gamma^3+4\gamma^2-\gamma-1}\left(\frac{d\Delta}{4n}\right)^{1-\frac{1}{2\gamma}}+\frac{\Delta}{7}\left(\frac{15\pi}{4}\right)^{2/5}\left(\frac{n}{d^2}\right)^{2/5}\\\mathif d\ll n\ll d^2 \mathand p\ge d\mathand\tilde{\lambda}\ll\frac{\sqrt{n}}{d}\end{aligned}.
    \end{equation}
    The rank of the ERM solution is also given by \eqref{eq:rank-ERM}, because the total number of spikes is subleading. Therefore, we can write
    \begin{equation}
    \boxed{\begin{aligned}\caE=\frac{24\gamma^3}{4\gamma^3+4\gamma^2-\gamma-1}\left(\frac{d\Delta}{4n}\right)^{1-\frac{1}{2\gamma}}+\frac{\Delta}{7}\left(\frac{15\pi}{4}\right)^{2/5}\left(\frac{n}{d^2}\right)^{2/5}\\\mathif d\ll n\ll d^2 \mathand p\gg n^{3/5}d^{-1/5}\mathand\tilde{\lambda}\ll\frac{\sqrt{n}}{d}\end{aligned}}.
    \label{eq:rank-not-constrain-error}
    \end{equation}
    This corresponds to the pink and the purple region on the right of the dashed line in Figure \ref{fig:fig2}(a).

    \item\textbf{Phase VI}: all spikes are recovered, thus the ERM solution has full rank, and  the excess error is given by
    \begin{equation}
    \boxed{\caE=\Theta\left(\frac{d^2}{n}\right)\mathif\tl\ll\sqrt{\frac{n}{d^2}}\mathand n\gg d^2\mathand p\geq d},
    \end{equation}
    or equivalently
    \begin{equation}
    \boxed{\beta=2-\alpha\mathif\ell<\frac{\alpha}{2}-1\mathand\alpha>2\mathand\rho\geq1}.
    \end{equation}
    This corresponds to the green region in Figure \ref{fig:fig2}(a).
\end{itemize}

\paragraph{Case I: rank constraint active, spikes only.}
We start by considering the case in which the final spectrum is composed by exactly $p$ spikes. 
This is the case if the $p$-th target eigenvalue $\sqrt{d} p^{-\gamma}$ is pushed out of the bulk by the BBP mapping, $\sqrt{d} p^{-\gamma} \gg \delta$. Once this $p$ eigenvalues are out, the rank-constraint will set to zero the remainder of the possibly pushed-out target eigenvalues, as well as the semicircular bulk. 
Additionally, the PSD+regularization constraint will set to zero all eigenvalues smaller than $\tl \epsilon$, requiring for the $p$-th spike to be non zero that $\sqrt{d}p^{-\gamma}+\frac{\delta^2}{\sqrt{d}p^{-\gamma}}-\tl\epsilon \gg 0$.
Overall, this implies $\sqrt{d} p^{-\gamma} \gg \max(\delta, \tl\epsilon)$.
Notice that here we used $\gg$ where a priori only $\geq$ was needed in both conditions: this is a simplifying assumption that would alter the final phase diagram Figure \ref{fig:fig2} only in a zero measure set.

We can then write $J_p$ as 
\begin{align}
        &J_p(\delta,\tl\epsilon) =\frac{1}{d}\sum_{i=1}^{p}\left[\sqrt{d}i^{-\gamma}+\frac{\delta^2}{\sqrt{d}i^{-\gamma}}-\tl\epsilon\right]^2\\
    &=\sum_{i=1}^{p}\left[i^{-2\gamma}+\frac{\delta^4}{d^2i^{-2\gamma}}+\frac{1}{d}\lambda^2\epsilon^2-2\tl\epsilon\frac{i^{-\gamma}}{\sqrt
    {d}}-2\tl\epsilon\frac{\delta^2}{d^{3/2}i^{-\gamma}}+2\frac{\delta^2}{d}\right]\\
    &\approx Q^\star-\frac{p^{1-2\gamma}}{2\gamma-1}+\frac{\delta^4}{d^2}\frac{p^{1+2\gamma}}{1+2\gamma}+\frac{p}{d}(\lambda^2\epsilon^2+2\delta^2)-\frac{2\tl\epsilon}{\sqrt{d}}\sum_{i=1}^{p}\left[i^{-\gamma}+\frac{\delta^2}{di^{-\gamma}}\right]
\end{align}
The derivatives with respect to $\tl\epsilon$ and $\delta$ are
\begin{align}
\begin{dcases}
    \tl\epsilon\del_2J_p(\delta,\tl\epsilon)\approx 2\frac{p}{d}\lambda^2\epsilon^2-\frac{2\tl\epsilon}{\sqrt{d}}\sum_{i=1}^{p}\left[i^{-\gamma}+\frac{\delta^2}{di^{-\gamma}}\right]\\
    \delta\del_1 J_p(\delta,\tl\epsilon)\approx 4\frac{\delta^4}{d^2}\frac{p^{1+2\gamma}}{1+2\gamma}+4\frac{p}{d}\delta^2-\frac{4\tl\epsilon\delta^2}{d^{3/2}}\sum_{i=1}^{p}i^\gamma
\end{dcases}
\end{align}
such that we can write \eqref{eq:SE_ERM} as
\begin{align}
&
    \begin{dcases}
        4\frac{n}{d^2}\delta^2-\frac{\delta^2}{\epsilon}\approx 4\frac{\delta^4}{d^2}\frac{p^{1+2\gamma}}{1+2\gamma}+4\frac{p}{d}\delta^2-\frac{4\tl\epsilon\delta^2}{d^{3/2}}\frac{p^{1+\gamma}}{1+\gamma}\ll \frac{p^{1-2\gamma}}{2\gamma-1}\\
        Q^\star+\frac{\Delta}{2}+2\frac{n}{d^2}\delta^2-\frac{\delta^2}{\epsilon}\approx Q^\star-\frac{p^{1-2\gamma}}{2\gamma-1}+\frac{\delta^4}{d^2}\frac{p^{1+2\gamma}}{1+2\gamma}+\frac{p}{d}(2\delta^2-\lambda^2\epsilon^2)\approx Q^\star-\frac{p^{1-2\gamma}}{2\gamma-1}
    \end{dcases}\\
    &\qquad\implies \caE(p)\approx \frac{p^{1-2\gamma}}{2\gamma-1} \approx Q^\star-\sum_{i=1}^{p}i^{-2\gamma},
\end{align}
where we used the conditions for having $p$ spikes visible  ($\sqrt{d}p^{-\gamma}\gg \max(\delta,\tl\epsilon)$) to compare the terms and keep only the leading ones ($\propto p^{1-2\gamma}$).
The first equation gives in this limit $\epsilon\approx \frac{d^2}{4n}$, while the second gives at zeroth order $\delta_0\approx \frac{d}{2}\sqrt{\frac{\Delta}{n}}$, using that $p\gg 1$. The condition $\max(\delta_0,\tl\epsilon)\ll \sqrt{d}p^{-\gamma}$ then translates to $n\gg \max{(\Delta dp^{2\gamma},\frac{1}{4}\lambda d^{3/2}p^\gamma)}$. Thus the excess error is
\begin{equation}
\boxed{
\caE=\frac{1}{2\gamma-1}p^{1-2\gamma}\mathif n\gg \max{(dp^{2\gamma},\lambda d^{3/2}p^\gamma)}.
}
\end{equation}
We can thus write this phase using the notation of the main text, also resolving the maximum, as
\begin{equation}
    \boxed{\beta=\rho(1-2\gamma)\mathif 
    \begin{dcases}
        \alpha>1+2\gamma\rho\mathand \ell<\frac{\alpha}{2}-1\\
        \alpha>\ell+\frac{3}{2}+\gamma\rho\mathand \ell>\frac{\alpha}{2}-1
    \end{dcases}}
\end{equation}
This corresponds to the white regions of Figure \ref{fig:fig2}.

\paragraph{Case II: rank constraint active, small bulk.}
We now consider the case in which the final spectrum is composed by $1 \ll K \ll p$ spikes, and by a slice of the semi-circular bulk. For this to be the case, we need to impose that:
\begin{itemize}
    \item The $p$-th spike is not pushed out of the bulk by the BBP mapping, i.e. $\sqrt{d} p^{-\gamma} \ll \delta$, while the first spike is, i.e. $\sqrt{d} \gg \delta$. The fact that just $K$ spikes are out imposes that $K^\gamma \approx \sqrt{d} / \delta$.
    \item We also need to impose that the rank cutoff is active, i.e. $x_0 > 0$ and that the overall cutoff at $x_0 + \tl \epsilon$ is inside the bulk, i.e. $x_0 + \tl\epsilon < 2\delta$. We will assume that $s:=2-\frac{x_0+\tl\epsilon}{\delta}\ll 1$.
    \item Finally, we need to impose that the remaining bulk is small compared with the non-width-constrained networks: $s\ll t:=2-\frac{\tl\epsilon}{\delta}$.
\end{itemize}
The spike contribution to $J_p$ reads
\begin{align}\label{eq:Jp2_overfitting}
    &J_p^{(2)}(\delta,\tl\epsilon) =\frac{1}{d}\sum_{i=1}^{K}\left[\sqrt{d}i^{-\gamma}+\frac{\delta^2}{\sqrt{d}i^{-\gamma}}-\tl\epsilon\right]^2\\
    &=\sum_{i=1}^{K}\left[i^{-2\gamma}+\frac{\delta^4}{d^2i^{-2\gamma}}+\frac{1}{d}\tl^2\epsilon^2-2\tl\epsilon\frac{i^{-\gamma}}{\sqrt
    {d}}-2\tl\epsilon\frac{\delta^2}{d^{3/2}i^{-\gamma}}+2\frac{\delta^2}{d}\right]\\
    &\approx Q^\star+\frac{K^{1-2\gamma}}{1-2\gamma}+\frac{K}{d}(\tl^2\epsilon^2+2\delta^2)+\frac{\delta^4}{d^2}\frac{K^{1+2\gamma}}{1+2\gamma}-\frac{2\tl\epsilon\delta^2}{d^{3/2}}\sum_{i=1}^{K}i^{\gamma}-\frac{2\tl\epsilon}{\sqrt{d}}\sum_{i=1}^{K}i^{-\gamma}
\end{align}
For the bulk piece, using \eqref{eq:bulk_J_approx} and $s\ll t$ to get the leading order in $s$, we have
\begin{align}
    J_p^{(1)}(\delta,\tl\epsilon)\approx  \frac{2\delta^2}{3\pi} t^2 s^{3/2} -  \frac{4\delta^2}{5\pi} t s^{5/2} +  \frac{2\delta^2}{7\pi}  s^{7/2}\approx{\frac{2\delta^2}{3\pi} t^2 s^{3/2}}\,,
\end{align}
Before computing the derivatives, we need to recall that $ K^\gamma\approx\frac{\sqrt{d}}{\delta}$. 
We also need to consider how $x_0$ depends on $p$ using \eqref{eq:bulk_int_approx} and \eqref{eq:cutoff}:
\begin{align}
    p=K+d\int_{\tl\epsilon+x_0}^{2\delta}\mu_{sc,\delta}(x)\dd x =K+d\int_{2-s}^2\mu_{sc}(x)\dd x\approx\frac{2d}{3\pi}s^{3/2},
\end{align}
where for the last approximation we use $s\ll 1$ and $K\ll p$.

We use these relations to rewrite $J_p$ and compute the derivatives
\begin{align}
    J_p(\delta,\tl\epsilon)&\approx Q^\star+\frac{K^{1-2\gamma}}{1-2\gamma}+\frac{K}{d}(\tl^2\epsilon^2+2\delta^2)+\frac{\delta^4}{d^2}\frac{K^{1+2\gamma}}{1+2\gamma}-\frac{2\tl\epsilon\delta^2}{d^{3/2}}\sum_{i=1}^{K}i^{\gamma}-\frac{2\tl\epsilon}{\sqrt{d}}\sum_{i=1}^{K}i^{-\gamma}+ \frac{p}{d} t^2\delta^2\\
    &\approx Q^\star+K^{1-2\gamma}\left[\frac{1}{1-2\gamma}+\frac{\tl^2\epsilon^2}{\delta^2}+2+\frac{1}{1+2\gamma}-\left(2-\frac{\tl\epsilon}{\delta}\right)^2\right]-\frac{2\tl\epsilon}{\delta}\frac{K^{1-2\gamma}}{1+\gamma}-\frac{2\tl\epsilon}{\delta}\frac{K^{1-2\gamma}}{1-\gamma}+ \frac{p}{d} t^2\delta^2\\
        &\approx Q^\star+\left(\frac{\delta}{\sqrt{d}}\right)^{2-\frac{1}{\gamma}}\left[\frac{2}{1-4\gamma^2}-2-\frac{4\tl\epsilon}{\delta}\frac{\gamma^2}{1-\gamma^2}\right]+ \frac{p}{d} (2\delta-\tl\epsilon)^2\\
    \implies &
        \tl\epsilon\del_2J_p(\delta,\tl\epsilon)\approx -\frac{4\tl\epsilon}{\delta}\frac{\gamma^2}{1-\gamma^2}\left(\frac{\delta}{\sqrt{d}}\right)^{2-\frac{1}{\gamma}}- \frac{2p}{d} \tl\epsilon(2\delta-\tl\epsilon)\\
      \mathand  &\delta\del_1J_p(\delta,\tl\epsilon)\approx \frac{2\gamma-1}{\gamma}\left(\frac{\delta}{\sqrt{d}}\right)^{2-\frac{1}{\gamma}}\left[\frac{2}{1-4\gamma^2}-2-\frac{4\tl\epsilon}{\delta}\frac{\gamma^2}{1-\gamma^2}\right]+\frac{4\tl\epsilon}{\delta}\frac{\gamma^2}{1-\gamma^2}\left(\frac{\delta}{\sqrt{d}}\right)^{2-\frac{1}{\gamma}}+ \frac{4p}{d} \delta(2\delta-\tl\epsilon)\\
        &\qquad= C\left(\gamma,t\right)\left(\frac{\delta}{\sqrt{d}}\right)^{2-\frac{1}{\gamma}}+ \frac{4p}{d} \delta^2t
\end{align}
where $C(\gamma,t) = O(1)$ is a constant that depends on the value of $t$ and $\gamma$. Now we can write the state evolution equations \eqref{eq:SE_ERM} as
\begin{align}
    &\begin{dcases}\label{eq:SE_rank_overfit}
        4\frac{n}{d^2}\delta^2-\frac{\delta^2}{\epsilon}\approx C\left(\gamma,t\right)\left(\frac{\delta}{\sqrt{d}}\right)^{2-\frac{1}{\gamma}}+  \frac{4p}{d} \delta^2 t\\
        Q^\star+\frac{\Delta}{2}+2\frac{n}{d^2}\delta^2-\frac{\delta^2}{\epsilon}\approx Q^\star+\left(\frac{\delta}{\sqrt{d}}\right)^{2-\frac{1}{\gamma}}\left[\frac{2}{1-4\gamma^2}-2\right]+\frac{4p}{d}\delta^2 t-\frac{p}{d}\delta^2t^2
    \end{dcases}\\
    &\implies \caE(p)\approx \frac{p}{d} \delta^2t^2 + \left(\frac{\delta}{\sqrt{d}}\right)^{2-\frac{1}{\gamma}}\left[C\left(\gamma,t\right)-\frac{2}{1-4\gamma^2}+2\right]
    \label{eq:error-caseII}
\end{align}
We now need to solve the state evolution equations. Notice that $K\gg1$ implies $\left(\frac{\delta}{\sqrt{d}}\right)^{2-\frac{1}{\gamma}}\ll1$. There are two cases.
\begin{itemize}
\item The first case is when $\frac{p}{d}\delta^2 t,\frac{p}{d}\delta^2t^2\ll1$. Then \eqref{eq:SE_rank_overfit} reduces to 
\begin{equation}
\begin{dcases}
    4\frac{n}{d^2}\delta^2-\frac{\delta^2}{\epsilon}\approx 0\\
    Q^\star+\frac{\Delta}{2}+2\frac{n}{d^2}\delta^2-\frac{\delta^2}{\epsilon}\approx0
\end{dcases},
\end{equation}
from which we obtain the leading order solution $\epsilon\approx\frac{d^2}{4n}, \delta^2\approx\frac{\Delta d^2}{4n}$. This gives the excess risk
\begin{equation}
\caE=\frac{\Delta pd}{n}\left(1-\frac{\tilde{\lambda}d}{4\sqrt{\Delta n}}\right)^2+\left[C\left(\gamma,t\right)-\frac{2}{1-4\gamma^2}+2\right]\left(\frac{\Delta d}{4n}\right)^{1-\frac{1}{2\gamma}},
\end{equation}
where we use $t:=2-\frac{\tl\epsilon}{\delta}=2-\frac{\tl d}{\sqrt{\Delta n}}$. The conditions defining the boundaries of this phase are
\begin{align}
    \begin{cases}
        t>0\\
        s\ll t\\
        1\ll K\ll p\\
        \frac{p}{d}\delta^2 t,\frac{p}{d}\delta^2t^2\ll1
    \end{cases}
    \implies
    \begin{cases}
        \tl<4\sqrt{\Delta}\frac{\sqrt{n}}{d}\\
        p\ll d\left(2-\frac{\tl d}{\sqrt{\Delta n}}\right)^{3/2}\\
        1\ll n\ll dp^{2\gamma}\\
        \frac{n}{pd}\gg 2-\frac{\tl d}{\sqrt{\Delta n}},\frac{pd}{n}\left(1-\frac{\tilde{\lambda}d}{4\sqrt{\Delta n}}\right)^2\ll1
        \label{eq:boundary-cases}
    \end{cases}
\end{align}
As a special case\footnote{In fact \eqref{eq:boundary-cases} also includes some interesting boundary cases, which is dropped for simplicity because they are not the focus of our paper.}, we can write
\begin{equation}
\boxed{\caE=\frac{\Delta pd}{n}\mathif pd\ll n\ll p^{2\gamma}d\mathand\tilde{\lambda}\ll\frac{\sqrt{n}}{d}}.
\end{equation}
Equivalently,
\begin{equation}
    \boxed{\beta=\rho-\alpha+1\mathif 1+2\gamma\rho>\alpha>1+\rho\mathand \ell<\frac{\alpha}{2}-1}
\end{equation}
This corresponds to the orange region of Figure \ref{fig:fig2}(a).

\item The second case is when $\epsilon\gg\frac{d^2}{n}$ and $\frac{p}{d}\delta^2t^2\ll\frac{p}{d}\delta^2t=\Theta(1)$. Then \eqref{eq:SE_rank_overfit} reduces to 
\begin{equation}
\begin{dcases}
    4\frac{n}{d^2}\delta^2\approx\frac{4p}{d}\delta^2t\\
    Q^\star+\frac{\Delta}{2}+2\frac{n}{d^2}\delta^2\approx Q^\star+\frac{4p}{d}\delta^2t
\end{dcases}.
\end{equation}
The leading order solution is thus $t\approx\frac{n}{pd}$ and $\delta^2\approx\frac{\Delta d^2}{4n}$ with $\epsilon=\frac{2\delta(1-t)}{\tl}$. Taking it into \eqref{eq:error-caseII}, we obtain
\begin{equation}
\caE=\frac{\Delta n}{4pd}+\left[C\left(\gamma,t\right)-\frac{2}{1-4\gamma^2}+2\right]\left(\frac{\Delta d}{4n}\right)^{1-\frac{1}{2\gamma}}.
\end{equation}
The conditions defining the boundaries of this phase are
\begin{align}
    \begin{cases}
        t>0\\
        s\ll t\\
        1\ll K\ll p\\
        \frac{p}{d}\delta^2 t\ll\frac{p}{d}\delta^2t^2=\Theta(1)\\
        \epsilon\gg\frac{d^2}{n}
    \end{cases}
    \implies
    \begin{cases}
        p\ll n^{3/5}d^{-1/5}\\
        1\ll n\ll dp^{2\gamma}\\
        n\ll pd\\
        \tl\ll\frac{\sqrt{n}}{d}
    \end{cases}
\end{align}
In conclusion, this phase can be written as 
\begin{equation}
\boxed{\begin{aligned}&\caE=\frac{\Delta n}{4pd}+\frac{24\gamma^3}{(4\gamma^2-1)(\gamma+1)}\left(\frac{\Delta d}{4n}\right)^{1-\frac{1}{2\gamma}}\\&\mathif d\ll n\ll pd \mathand p\ll n^{3/5}d^{-1/5}\mathand\tilde{\lambda}\ll\frac{\sqrt{n}}{d}\end{aligned}},
\label{eq:rank-constrain-error}
\end{equation}
where the constant $\frac{24\gamma^3}{(4\gamma^2-1)(\gamma+1)}$ is obtained by $C(\gamma,0)=\frac{8\gamma^2}{(\gamma+1)(2\gamma+1)}$ given that $t\ll1$ in the phase.
This corresponds to the red and purple regions of Figure \ref{fig:fig2}(a).
We stress that this covers only the purple region on the right side of the dashed line, after which the rank constraint is not active anymore.

Combining \eqref{eq:rank-constrain-error} and \eqref{eq:rank-not-constrain-error}, we can obtain the following three cases:
\begin{equation}
    \boxed{\beta=\alpha-\rho-1\mathif 1+\rho>\alpha>\max\left(1+\rho\frac{2\gamma}{4\gamma-1},\frac{5\rho+1}{3}\right)\mathand \ell<\frac{\alpha}{2}-1},
\end{equation}
\begin{equation}
    \boxed{\beta=2(\alpha-2)/5\mathif \frac{5\rho+1}{3}>\alpha>\frac{18\gamma-5}{14\gamma-5}\mathand \ell<\frac{\alpha}{2}-1},
\end{equation}
and
\begin{equation}
\boxed{\beta=-(\alpha-1)\left(1-\frac{1}{2\gamma}\right)\mathif \min\left(\frac{18\gamma-5}{14\gamma-5},1+\rho\frac{2\gamma}{4\gamma-1}\right)>\alpha>1\mathand \ell<\frac{\alpha}{2}-1}.
\end{equation}
\item The case in which  $\frac{p}{d}\delta^2 t,\frac{p}{d}\delta^2t^2\gg1$ is treated later when considering the interpolation peak.
\end{itemize}

\paragraph{Case III: rank constraint active, no spikes.} In this regime, the spectrum of $\bhS$ is without spikes $K=0$ ($\delta\gg\sqrt{d}$), thus we have
\begin{equation}
    \mu_\delta(x) \approx \mu_{\rm sc}(x/\delta) / \delta \,,
\end{equation}
The moment $J_p$ becomes
\begin{equation}
J_p(\delta,\tl\epsilon)\approx\int_{x_0^+}^{2\delta - \tl\epsilon}x^2\mu_{\rm s.c.}\big((x + \tl \epsilon)/\delta\big) / \delta \,{\rm d} x = \delta^2\int_{x_0^+/\delta+\tl\epsilon/\delta}^{2}(x-\tl\epsilon/\delta)^2\mu_{\rm s.c.}(x) \,{\rm d} x\,.
\end{equation}
Calling $t  :=2- \tl\epsilon/\delta$ and $s  := t  - x_0^+/\delta$, we do the scaling ansatz $s  \ll t \ll 1$ (i.e. the rank constraint dominates the spectral cut). This gives
\begin{equation}
    J_p(\delta,\tl\epsilon)\approx \frac{2\delta^2}{3\pi}t ^2 s ^{3/2}\,.
\end{equation}
Using the definition of the cutoff 
\eqref{eq:cutoff}
we get
\begin{equation}
\frac{p}{d} = \int_{\tl\epsilon + x_0^+}^{2\delta} \mu_\delta(x)\,{\rm d} x\, \approx \frac{2}{3\pi} s ^{3/2} \,,
\end{equation}
which gives
\begin{equation}
J_p(\delta,\tl\epsilon)\approx \frac{p}{d}t ^2 \delta^2.
\end{equation}
The derivatives of $J_p$ include many terms, but the leading ones for $t \ll1$
are given by
\begin{equation}
\partial_1J_p(\delta,\tl\epsilon)\approx 4\frac{p}{d}t  \delta\,,
\mathand
\tl\epsilon\partial_2J_p(\delta,\tl\epsilon)\approx -4\frac{p}{d}t  \delta^2\,,
\end{equation}
where we further used that $\tl\epsilon\approx2\delta$ (due to $t  \ll 1$). Plugging this into \eqref{eq:SE_ERM} we obtain
\begin{equation}
    \begin{cases}
    \begin{aligned}
        \displaystyle
        &\frac{4n}{d^2}\,\delta^2 - \frac{\delta^2}{\epsilon}
        \approx
        4\frac{p}{d}t  \delta^2,
        \\
        \displaystyle
        &Q^* + \frac{\Delta}{2}
        + \frac{2n}{d^2}\,\delta^2
        - \frac{\delta^2}{\epsilon}
        \approx 4\frac{p}{d}t  \delta^2,
    \end{aligned}
    \end{cases}
\label{eq:SE_ERM_case1}
\end{equation}
where we use $t \ll1$ again to state that the term $J_p$ is much smaller than its derivatives. The first equation of \eqref{eq:SE_ERM_case1} subtracted by the second yields the prediction for the test error
\begin{equation}
\caE:=2\frac{n}{d^2}\delta^2-\frac{\Delta}{2}\approx Q^* \, ,
\end{equation}
which is the same value of error as if $n=0$. Solving for $\delta$ we get
$\delta\approx\sqrt{(\Delta/2+Q^*)d^2/n}$.
To solve for $t $, we make another ansatz: $\epsilon\gg d^2/n$. Plugging this into the first equation of \eqref{eq:SE_ERM_case1} 
\begin{equation}
\frac{4n}{d^2}\delta^2\approx 4\frac{p}{d}t \delta^2,
\end{equation}
we can solve for $t $, 
obtaining $t \approx n / (pd)$.

The various assumptions that we made in the derivation lead to the following boundaries
\begin{align}
    \begin{cases}
        s\ll t\ll1\\
        \epsilon\gg\frac{d^2}{n}\\
        \delta\gg\sqrt{d}
    \end{cases}
    \implies
    \begin{cases}
      p\gg n^{3/5}d^{-1/5}\\
      \tl\ll\sqrt{\frac{n}{d^2}}\\
      n\ll d.
    \end{cases}
\end{align}
This corresponds to the yellow region on Figure \ref{fig:fig2}(a) on the left side of the dashed line.

Combining it with \eqref{eq:phaseIa}, we obtain
\begin{equation}
\boxed{
    \caE=Q^\star \mathif \tl\ll\sqrt{\frac{n}{d^2}}\mathand n\ll d},
\end{equation}
or equivalently
\begin{equation}
\boxed{\beta=0\mathif\alpha<1\mathand\ell>\frac{\alpha}{2}-1}.
\end{equation}

\paragraph{Case IV: Interpolation peak.}
We start from \eqref{eq:SE_rank_overfit} and \eqref{eq:error-caseII} with the ansatz that $\frac{p}{d}\delta^2t,\frac{p}{d}\delta^2t^2\gg1$:
\begin{align}
    &\begin{dcases}
    \label{eq:interpolate-SE}
        4\frac{n}{d^2}\delta^2-\frac{\delta^2}{\epsilon}\approx \frac{4p}{d} \delta^2 t\\
        Q^\star+\frac{\Delta}{2}+2\frac{n}{d^2}\delta^2-\frac{\delta^2}{\epsilon}\approx Q^\star+\frac{4p}{d}\delta^2 t-\frac{p}{d}\delta^2t^2
    \end{dcases}\\
    &\implies \caE(p)\approx \frac{p}{d} \delta^2t^2,
\end{align}
where we use the fact that $\left(\frac{\delta}{\sqrt{d}}\right)^{2-\frac{1}{\gamma}}\ll1$ as long as $K\gg1$. \eqref{eq:interpolate-SE} can be rewritten as
\begin{align}
\begin{cases}
&4\frac{n}{d^2}\delta^2-\frac{\delta^2}{\epsilon}=8\frac{p}{d}\delta^2-\frac{4p}{d}\tilde\lambda\epsilon\delta\\
&Q_\star +\frac{\Delta}{2}+2\frac{n}{d^2}\delta^2-\frac{\delta^2}{\epsilon}=Q_\star +\frac{4p}{d}\delta^2-\frac{p}{d}\tl^2\epsilon^2.
\end{cases}
\end{align}
Assuming $n=2pd$ and $\tl\epsilon\ll\delta$ (i.e. $t\approx2$), we can find the solution
\begin{align}
    \begin{dcases}
    \delta\approx \left(\frac{\Delta^2d}{4p\tl}\right)^{1/3},~\epsilon\approx \frac{\delta^2}{\Delta}\approx \left(\frac{d\sqrt{\Delta}}{4p\tl}\right)^{2/3}\\
     \caE\approx\frac{4p}{d}\delta^2\approx \tl^{-2/3}\left(\frac{4p\Delta^{4}}{d}\right)^{1/3}
     \end{dcases}
     \end{align}
The conditions defining the boundaries of this phase are
\begin{align}
    \begin{cases}
        n=2pd\\
        s\ll1\\
        \frac{p}{d}\delta^2t,\frac{p}{d}\delta^2t^2\gg1\\
        1\ll K\ll p
    \end{cases}
    \implies
    \begin{cases}
      n=2pd\\
      p\ll d\\
      \frac{1}{p\sqrt{d}}\ll\tl\ll\sqrt{\frac{p}{d}}.
    \end{cases}
\end{align}
In this phase the excess error then reads
\begin{equation}
\boxed{\caE=2\left(\frac{d\Delta^2}{16p}\right)^{2/3}\tilde\lambda^{-2/3}\mathif n=2pd\mathand \frac{1}{p\sqrt{d}}\ll\tl\ll\sqrt{\frac{p}{d}}\mathand p\ll d}.
\label{eq:peak1}
\end{equation}
When $\tl\ll\frac{1}{p\sqrt{d}}$, $K\gg1$ is not satisfied and we have no spikes outside the bulk so the contribution from $J_p^{(1)}$ vanishes. Using the same assumptions on the bulk part as in \eqref{eq:SE_rank_overfit}, we get instead
\begin{align}
\begin{cases}
&4\frac{n}{d^2}\delta^2-\frac{\delta^2}{\epsilon}=8\frac{p}{d}\delta^2-\frac{4p}{d}\tilde\lambda\epsilon\delta\\
&Q_\star +\frac{\Delta}{2}+2\frac{n}{d^2}\delta^2-\frac{\delta^2}{\epsilon}=\frac{4p}{d}\delta^2-\frac{p}{d}\tl^2\epsilon^2.
\end{cases}
\end{align}
from which we can easily see that the only difference is be the constant (since the constant $\frac{\Delta}{2}$ is replaced with $Q^\star+\frac{\Delta}{2}$ in the equations), which gives
\begin{equation}
\boxed{\caE=2\left(\frac{d(\Delta+2Q^\star)^2)}{16p}\right)^{2/3}\tilde\lambda^{-2/3}\mathif n=2pd\mathand \tl\ll\frac{1}{p\sqrt{d}}\mathand p\ll d}.
\label{eq:peak2}
\end{equation}
Combining \eqref{eq:peak1} and \eqref{eq:peak2} we obtain the interpolation peak:
\begin{equation}
    \boxed{\beta=-\frac{2}{3}\ell+\frac{\rho-1}{3}\mathfor\ell<\frac{\alpha}{2}-1\mathand \alpha=\rho+1\mathand\rho<1}
\end{equation}

\subsection{Derivation of the excess test error for width-constrained, noiseless networks}
\label{app:noiseless}

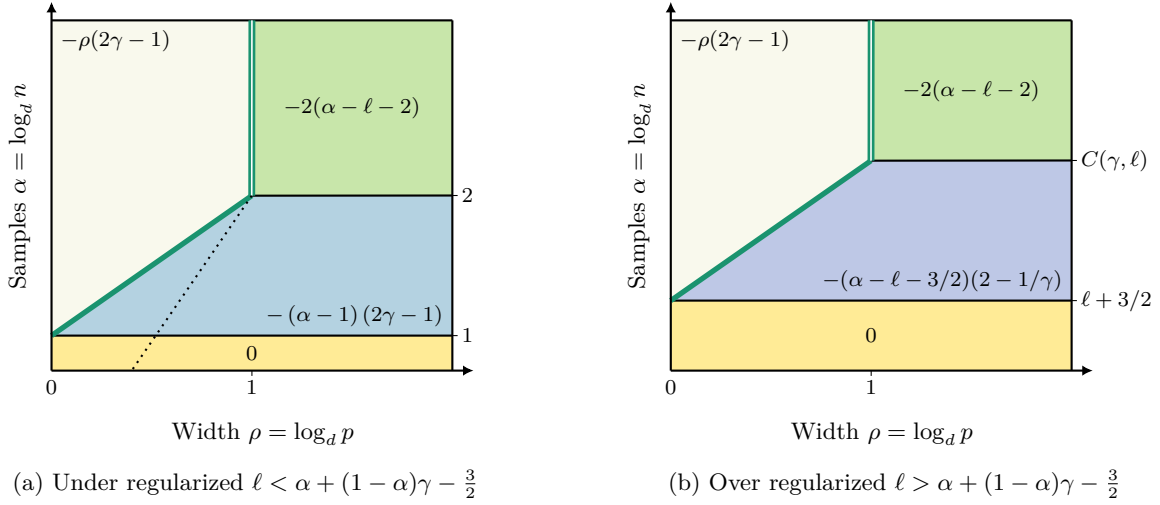
\begin{figure}[t]
\centering
\begin{subfigure}[t]{0.48\linewidth}
    \centering
    \begin{tikzpicture}[scale=13.96]

\tikzstyle{every node}=[font=\small]



\definecolor{bulk}{RGB}{210,210,210}

\definecolor{phaseI}{RGB}{255,235,150}
\definecolor{phaseII}{RGB}{180,210,225}
\definecolor{phaseIII}{RGB}{200,230,170}
\definecolor{phaseIV}{RGB}{190,200,230}
\definecolor{phaseV}{RGB}{255,200,170}
\definecolor{phaseVI}{RGB}{170,235,210}

\definecolor{phaseLow}{RGB}{249,249,237}
\definecolor{mygreen}{RGB}{26, 147, 111}    

\def\Ox{0.08}
\def\Oy{0.09}
\def\AXx{0.40}
\def\AXy{0.35}
\def\height{0.95*\AXy}
\def\width{0.95*\AXx}

\def\overparametrixed{0.5*\width}
\def\plateau{0.2*\width}

\def\linear{0.1*\height}
\def\peak{0.25*\height}
\def\quadratic{0.5*\height}
\def\extreme{0.9*\height}

\coordinate (origin) at (\Ox, \Oy);
\coordinate (LYtop) at (\Ox, \Oy + \height);
\coordinate (RYtop) at (\Ox + \width, \Oy + \height);
\coordinate (RYbottom) at (\Ox + \width, \Oy);

\coordinate (LY1) at (\Ox, \Oy + \linear);
\coordinate (LY2) at (\Ox, \Oy + \peak);
\coordinate (LY3) at (\Ox, \Oy + \quadratic);
\coordinate (LY4) at (\Ox, \Oy + \extreme);

\coordinate (RY1) at (\Ox + \width, \Oy + \linear);
\coordinate (RY2) at (\Ox + \width, \Oy + \peak);
\coordinate (RY3) at (\Ox + \width, \Oy + \quadratic);
\coordinate (RY4) at (\Ox + \width, \Oy + \extreme);

\coordinate (BX1) at (\Ox + \plateau, \Oy);
\coordinate (BX2) at (\Ox + \overparametrixed, \Oy);

\coordinate (TX2) at (\Ox + \overparametrixed, \Oy + \height);

\path[name path=OVER] (BX2) -- (TX2);

\path[name path=LINEAR]     (LY1) -- (RY1);
\path[name path=PEAK]       (LY2) -- (RY2);
\path[name path=QUADRATIC]  (LY3) -- (RY3);
\path[name path=EXTREME]    (LY4) -- (RY4);

\path[name intersections={of=OVER and EXTREME, by=O2}];
\path[name intersections={of=OVER and QUADRATIC, by=O1}];

\path[name path=OBLIQUE]    (BX1) -- (O1);
\path[name intersections={of=OBLIQUE and PEAK, by=P}];

\path[name path=PLATEAU]     (LY1) -- (O2);
\path[name intersections={of=PLATEAU and QUADRATIC, by=T2}];
\path[name intersections={of=PLATEAU and PEAK, by=T1}];

\coordinate (PX) at (P |- origin);

\fill[phaseII!100]
  (LY1) -- (O1)  -- (RY3) -- (RYbottom) -- cycle;
\node[anchor=south east] (r) at (RY1) {\footnotesize $-\left(\alpha-1\right)(2\gamma-1)$};

\fill[phaseIII!100]
  (O1) -- (RY3) -- (RYtop)  -- (TX2) -- cycle;
\node at ($(O1)!0.5!(RYtop)$) {\footnotesize $-2(\alpha-\ell-2)$};

\fill[phaseLow!100]
  (LY1) -- (O1) -- (TX2)  -- (LYtop) -- cycle;
\node[anchor=north west] (r) at (LYtop) {\footnotesize $-\rho(2\gamma-1)$};

\fill[phaseI!100]
  (origin) -- (RYbottom) -- (RY1) -- (LY1) -- cycle;
\node at (barycentric cs:origin=1,RYbottom=1,RY1=1,LY1=1) {\footnotesize $0$};

\draw[black, thick] (LYtop) -- (RYtop);
\draw[black, thick] (RYtop) -- (RYbottom);

\draw[black, thick] (O1) -- (RY3);
\draw[black, thick] (LY1) -- (RY1);

\draw[mygreen, line width=2pt] (O1) -- (LY1);
\draw[mygreen, line width=1pt, double] (O1) -- (TX2);

\draw[black, thick, dotted] (O1) -- (BX1);


\draw[->,thick] (origin) -- (\Ox + \AXx,\Oy);
\draw[->,thick] (\Ox,\Oy) -- (\Ox,\Oy + \AXy);

\node[anchor=north] (r) at (\Ox + \AXx/2,\Oy-0.04) {Width $\rho = \log_d p$};
\node[anchor=south, rotate=90] (r) at (\Ox-0.01,\Oy + \AXy/2) {Samples $\alpha = \log_d n$};

\node[anchor=north] (r) at (origin) {\footnotesize $0$};

\draw (BX2) -- ($(BX2)+(0,-0.005)$);
\node[anchor=north] (r) at (BX2) {\footnotesize $1$};

\draw (RY3) -- ($(RY3)+(0.005,0)$);
\node[anchor=west, rotate=0] (r) at (RY3) {\footnotesize $2$};

\draw (RY1) -- ($(RY1)+(0.005,0)$);
\node[rotate=0, anchor=west] (r) at (RY1) {\footnotesize $1$};







\end{tikzpicture}
    \caption{Under regularized $\ell < \alpha+(1-\alpha)\gamma-\frac{3}{2}$}
\end{subfigure}
\hfill
\begin{subfigure}[t]{0.48\linewidth}
    \centering
    \begin{tikzpicture}[scale=13.96]

\tikzstyle{every node}=[font=\small]



\definecolor{bulk}{RGB}{210,210,210}

\definecolor{phaseI}{RGB}{255,235,150}
\definecolor{phaseII}{RGB}{180,210,225}
\definecolor{phaseIII}{RGB}{200,230,170}
\definecolor{phaseIV}{RGB}{190,200,230}
\definecolor{phaseV}{RGB}{255,200,170}
\definecolor{phaseVI}{RGB}{170,235,210}

\definecolor{phaseLow}{RGB}{249,249,237}
\definecolor{mygreen}{RGB}{26, 147, 111}    

\def\Ox{0.08}
\def\Oy{0.09}
\def\AXx{0.40}
\def\AXy{0.35}
\def\height{0.95*\AXy}
\def\width{0.95*\AXx}

\def\overparametrixed{0.5*\width}
\def\plateau{0.4*\width}

\def\linear{0.2*\height}
\def\peak{0.25*\height}
\def\quadratic{0.6*\height}
\def\extreme{0.9*\height}

\coordinate (origin) at (\Ox, \Oy);
\coordinate (LYtop) at (\Ox, \Oy + \height);
\coordinate (RYtop) at (\Ox + \width, \Oy + \height);
\coordinate (RYbottom) at (\Ox + \width, \Oy);

\coordinate (LY1) at (\Ox, \Oy + \linear);
\coordinate (LY2) at (\Ox, \Oy + \peak);
\coordinate (LY3) at (\Ox, \Oy + \quadratic);
\coordinate (LY4) at (\Ox, \Oy + \extreme);

\coordinate (RY1) at (\Ox + \width, \Oy + \linear);
\coordinate (RY2) at (\Ox + \width, \Oy + \peak);
\coordinate (RY3) at (\Ox + \width, \Oy + \quadratic);
\coordinate (RY4) at (\Ox + \width, \Oy + \extreme);

\coordinate (BX1) at (\Ox + \plateau, \Oy);
\coordinate (BX2) at (\Ox + \overparametrixed, \Oy);

\coordinate (TX2) at (\Ox + \overparametrixed, \Oy + \height);

\path[name path=OVER] (BX2) -- (TX2);

\path[name path=LINEAR]     (LY1) -- (RY1);
\path[name path=PEAK]       (LY2) -- (RY2);
\path[name path=QUADRATIC]  (LY3) -- (RY3);
\path[name path=EXTREME]    (LY4) -- (RY4);

\path[name intersections={of=OVER and EXTREME, by=O2}];
\path[name intersections={of=OVER and QUADRATIC, by=O1}];

\path[name path=OBLIQUE]    (BX1) -- (O1);
\path[name intersections={of=OBLIQUE and PEAK, by=P}];

\path[name path=PLATEAU]     (LY1) -- (O2);
\path[name intersections={of=PLATEAU and QUADRATIC, by=T2}];
\path[name intersections={of=PLATEAU and PEAK, by=T1}];

\coordinate (PX) at (P |- origin);

\fill[phaseIV!100]
  (LY1) -- (O1)  -- (RY3) -- (RYbottom) -- cycle;
\node[anchor=south east] (r) at (RY1) {\footnotesize $-(\alpha-\ell-3/2)(2-1/\gamma)$};

\fill[phaseIII!100]
  (O1) -- (RY3) -- (RYtop)  -- (TX2) -- cycle;
\node at ($(O1)!0.5!(RYtop)$) {\footnotesize $-2(\alpha-\ell-2)$};

\fill[phaseLow!100]
  (LY1) -- (O1) -- (TX2)  -- (LYtop) -- cycle;
\node[anchor=north west] (r) at (LYtop) {\footnotesize $-\rho(2\gamma-1)$};

\fill[phaseI!100]
  (origin) -- (RYbottom) -- (RY1) -- (LY1) -- cycle;
\node at (barycentric cs:origin=1,RYbottom=1,RY1=1,LY1=1) {\footnotesize $0$};

\draw[black, thick] (LYtop) -- (RYtop);
\draw[black, thick] (RYtop) -- (RYbottom);

\draw[black, thick] (O1) -- (RY3);
\draw[black, thick] (LY1) -- (RY1);

\draw[mygreen, line width=2pt] (O1) -- (LY1);
\draw[mygreen, line width=1pt, double] (O1) -- (TX2);


\draw[->,thick] (origin) -- (\Ox + \AXx,\Oy);
\draw[->,thick] (\Ox,\Oy) -- (\Ox,\Oy + \AXy);

\node[anchor=north] (r) at (\Ox + \AXx/2,\Oy-0.04) {Width $\rho = \log_d p$};
\node[anchor=south, rotate=90] (r) at (\Ox-0.01,\Oy + \AXy/2) {Samples $\alpha = \log_d n$};

\node[anchor=north] (r) at (origin) {\footnotesize $0$};

\draw (BX2) -- ($(BX2)+(0,-0.005)$);
\node[anchor=north] (r) at (BX2) {\footnotesize $1$};

\draw (RY3) -- ($(RY3)+(0.005,0)$);
\node[anchor=west, rotate=0] (r) at (RY3) {\footnotesize $C(\gamma,\ell)$};

\draw (RY1) -- ($(RY1)+(0.005,0)$);
\node[rotate=0, anchor=west] (r) at (RY1) {\footnotesize $\ell+3/2$};






\end{tikzpicture}
    \caption{Over regularized $\ell>\alpha+(1-\alpha)\gamma-\frac{3}{2}$}
\end{subfigure}
\caption{Phase diagram of the noiseless estimation task. The dashed line and the blue lines represent the same meaning as in Figure \ref{fig:fig2}. See Figure \ref{fig:fignoiseless} for numerical verifications.}
\label{fig:noiseless_low_rank}
\end{figure}

We first recall the results in \cite{defilippis2026scaling}.
\begin{itemize}
\item \textbf{Phase I}: Phase Ia, Ib and Case III in Appendix \ref{app:derivation_scaling} also holds for $\Delta=0$, which gives:
\begin{align}
\boxed{\mathcal{E}=Q^\star , \mathif n\ll \max(d,\tl d^{3/2})},
\end{align}
or equivalently
\begin{equation}
\boxed{\beta=0\mathif\alpha<\min\left(1,\ell+\frac{3}{2}\right)}.
\end{equation}
This corresponds to the yellow region in Figure \ref{fig:noiseless_low_rank}.

\item \textbf{Phase II}: the excess error is given by
\begin{equation}
\caE=\frac{2\gamma}{2\gamma-1}\left(\tl\frac{d^{3/2}}{4n}\right)^{\frac{2\gamma-1}{\gamma}}\mathif \max\left(\frac{d^{\gamma-3/2}}{n^{\gamma-1}},\frac{n}{d^{\gamma+\frac{3}{2}}}\right)\ll\tl\ll\frac{d^{3/2}}{n}\mathand p\geq d.
\end{equation}
Moreover, there are $K\approx(4n/\tl d^{3/2})^{1/\gamma}$. The error is the same if $p\gg K\implies\tl\gg\frac{n}{d^{3/2}p^\gamma}$, which gives
\begin{equation}\
\boxed{\caE=\frac{2\gamma}{2\gamma-1}\left(\tl\frac{d^{3/2}}{4n}\right)^{\frac{2\gamma-1}{\gamma}}\mathif \max\left(\frac{d^{\gamma-3/2}}{n^{\gamma-1}},\frac{n}{d^{\gamma+\frac{3}{2}}},\frac{n}{d^{\frac{3}{2}}p^\gamma}\right)\ll\tl\ll\frac{d^{3/2}}{n}},
\end{equation}
or equivalently
\begin{equation}
\boxed{\beta=-\left(\alpha-\ell-\frac{3}{2}\right)\left(2-\frac{1}{\gamma}\right)\mathif\ell>\alpha+(1-\alpha)\gamma-\frac{3}{2}\mathand\ell+\frac{3}{2}<\alpha<\ell+\frac{3}{2}+\gamma\max(\rho,1)}.
\end{equation}
This corresponds to the purple region in Figure \ref{fig:noiseless_low_rank}(b).

\item\textbf{Phase III}: the ERM solution has full rank and the excess error is given by
\begin{equation}
\boxed{\caE=\frac{\tl^2d^4}{16n^2}\mathif\tl\ll\frac{n}{d^{\gamma+3/2}}\mathand n\gg d^2\mathand p\geq d},
\end{equation}
or equivalently
\begin{equation}
\boxed{\beta=-2(\alpha-\ell-2)\mathif\alpha>\max(2,\ell+\gamma+\frac{3}{2})\mathand\rho\geq1}.
\end{equation}
This corresponds to the green regionsin Figure \ref{fig:noiseless_low_rank}.

\item\textbf{Phase IV}: the excess error is given by
\begin{equation}
\caE=2^{-2\gamma+1}\left(\frac{24\gamma^3}{(4\gamma^2-1)(\gamma+1)}\right)^{2\gamma}\left(\frac{d}{n}\right)^{2\gamma-1}\mathif \tl\ll\frac{d^{\gamma-3/2}}{n^{\gamma-1}}\mathand d\ll n\ll d^2\mathand p\geq d.
\end{equation}
Because the rank of the ERM solution is still \eqref{eq:rank-ERM}, we can write 
\begin{equation}
\caE=2^{-2\gamma+1}\left(\frac{24\gamma^3}{(4\gamma^2-1)(\gamma+1)}\right)^{2\gamma}\left(\frac{d}{n}\right)^{2\gamma-1}\mathif p\gg n^{3/5}d^{-1/5}.
\label{eq:noiseless-phaseIV}
\end{equation}
This corresponds to the blue region of Figure \ref{fig:noiseless_low_rank}(a) on the right side of the dashed line.
\end{itemize}

\paragraph{Case I: rank constraint active, spikes only}
The analysis of Case I in Appendix \ref{app:derivation_scaling} holds also for $\Delta=0$, which gives
\begin{equation}
    \begin{cases}
    \begin{aligned}
        \displaystyle
        &\frac{4n}{d^2}\,\delta^2 - \frac{\delta^2}{\epsilon}
        \approx0,
        \\
        \displaystyle
        &Q^\star
        + \frac{2n}{d^2}\,\delta^2
        - \frac{\delta^2}{\epsilon}
        \approx Q_\star-\frac{1}{2\gamma-1}p^{1-2\gamma}.
    \end{aligned}
    \end{cases}
    \implies  
    \begin{dcases}
    \caE:=2\frac{n}{d^2}\delta^2\approx \frac{1}{2\gamma-1}p^{1-2\gamma}\\
        \delta\approx\sqrt{\frac{p^{1-2\gamma} d^2}{2(2\gamma-1)n}},~\epsilon\approx \frac{d^2}{4n}
    \end{dcases}
\label{eq:SE_ERM_case2,noiseless}
\end{equation}
Since the leading order solution for $\delta$ is different, we need to compute the new boundaries from the conditions we assumed on the spectrum
\begin{align}
    \begin{dcases}
        \sqrt{d}p^{-\gamma}\gg \tl\epsilon+x_0^+>2\delta\\
        \sqrt{d}p^{-\gamma}\gg \max(2\delta,\tl\epsilon)
    \end{dcases}
    \implies
    \begin{dcases}
        n\gg pd \mathif \tl\ll \frac{p^{\frac{1}{2}-\gamma}\sqrt{n}}{d}\\
        n\gg \tl d^{3/2}p^\gamma \mathif  \tl\gg \frac{p^{\frac{1}{2}-\gamma}\sqrt{n}}{d}
    \end{dcases}
\end{align}
One can rewrite the regularization boundary to make it independent of $p$ by using that at the boundary $n\approx pd\approx \tl d^{3/2}p^\gamma\iff \alpha=1+\rho=\ell+\frac{3}{2}+\rho\gamma$, this gives the boundary $\ell_c:=\alpha(1-\gamma)+\gamma-\frac{3}{2}$. We can thus write this phase as 
\begin{equation}
\boxed{\caE=\frac{1}{2\gamma-1}p^{1-2\gamma}\mathif \tl\ll\frac{n}{p^\gamma d^{3/2}}\mathand n\gg pd},
\end{equation}
or equivalently
\begin{equation}
    \boxed{\beta=-\rho(2\gamma-1)\mathif \begin{dcases}
        \alpha>1+\rho \mathand \ell<\alpha(1-\gamma)+\gamma-\frac{3}{2}\\
        \alpha>\ell+\frac{3}{2}+\rho\gamma\mathand \ell>\alpha(1-\gamma)+\gamma-\frac{3}{2}
    \end{dcases}}.
\end{equation}
This corresponds to the white region in Figure \ref{fig:noiseless_low_rank}.

\paragraph{Case II: rank constraint active, before interpolation.} Case II corresponds to \eqref{eq:SE_rank_overfit} with 
$\Delta=0$. We make an ansatz that $\frac{p}{d}\delta^2t^2\ll (\delta/\sqrt{d})^{2-1/\gamma}$, and then from \eqref{eq:SE_rank_overfit} we approximately have
\begin{align}
\begin{cases}
&4\frac{n}{d^2}\delta^2-\frac{\delta^2}{\epsilon}=4\frac{p}{d}\delta^2t+\left(C_1(\gamma)(2-1/\gamma)+C_2(\gamma)\right)\left(\frac{\delta}{\sqrt{d}}\right)^{2-\frac{1}{\gamma}}\\
&Q^\star +2\frac{n}{d^2}\delta^2-\frac{\delta^2}{\epsilon}=Q^\star +4\frac{p}{d}\delta^2t +\left(C_1(\gamma)+C_2(\gamma)\right)\left(\frac{\delta}{\sqrt{d}}\right)^{2-\frac{1}{\gamma}},
\end{cases}
\label{eq:SE-noiseless,case IV}
\end{align}
for constants $C_1(\gamma):=-\frac{1}{2\gamma-1}+6-\frac{4}{1-\gamma}-\frac{4}{1+\gamma}+\frac{1}{1+2\gamma}$, $C_2(\gamma):=-4+\frac{4}{1-\gamma}+\frac{4}{1+\gamma}$. We make another ansatz that $\frac{1}{\epsilon}\ll\frac{n}{d^2}$. The difference between the two equations in \eqref{eq:SE-noiseless,case IV} give
\begin{align}
2\frac{n}{d^2}\delta^2=\frac{24\gamma^3}{(4\gamma^2-1)(\gamma+1)}\left(\frac{\delta}{\sqrt{d}}\right)^{2-\frac{1}{\gamma}}.
\label{eq:noiseless-caseII}
\end{align}
Taking it back to \eqref{eq:SE-noiseless,case IV}, we obtain the leading order solution $t =\frac{cn}{pd}$ for some constant $c$ depending on $\gamma$. We can solve $\delta$ from \eqref{eq:noiseless-caseII}:
\begin{align}
\delta=2^{-\gamma}\left(\frac{24\gamma^3}{(4\gamma^2-1)(\gamma+1)}\right)^\gamma\frac{d^{\gamma+1/2}}{n^\gamma},
\end{align}
which gives
\begin{align}
\caE:=2\frac{n}{d^2}\delta^2=2^{-2\gamma+1}\left(\frac{24\gamma^3}{(4\gamma^2-1)(\gamma+1)}\right)^{2\gamma}\left(\frac{d}{n}\right)^{2\gamma-1}.
\end{align}
$\epsilon$ is given by $\epsilon\approx\frac{2\delta}{\tl}$. The conditions defining the boundaries of this phase are
\begin{align}
    \begin{cases}
        \frac{p}{d}\delta^2t^2\ll(\delta/\sqrt{d})^{2-1/\gamma}\\
        s\ll t\ll1\\
        \frac{1}{\epsilon}\ll\frac{n}{d^2}\\
        \delta\ll\sqrt{d}
    \end{cases}
    \implies
    \begin{cases}
    p\ll n^{3/5}d^{-1/5}\\
    d\ll n\ll pd  \\
    \tl\ll\frac{d^{\gamma-3/2}}{n^{\gamma-1}}
    \end{cases}
\end{align}
This corresponds to the blue region of Figure \ref{fig:noiseless_low_rank}(a) on the left side of the dashed line.
Combining it with \eqref{eq:noiseless-phaseIV}, the excess error reads
\begin{equation}
\boxed{\caE=2^{-2\gamma+1}\left(\frac{24\gamma^3}{(4\gamma^2-1)(\gamma+1)}\right)^{2\gamma}\left(\frac{d}{n}\right)^{2\gamma-1}\mathif d\ll n\ll d^2 \mathand\tl\ll\frac{d^{\gamma-3/2}}{n^{\gamma-1}}\mathand p\gg\frac{n}{d},}
\end{equation}
or equivalently
\begin{equation}
\boxed{\beta=-(\alpha-1)(2\gamma-1)\mathif 1<\alpha<2\mathand\ell<\alpha+(1-\alpha)\gamma-\frac{3}{2}\mathand\rho>\alpha-1}.
\end{equation}

\paragraph{Summary} The phase diagram is shown in Figure \ref{fig:noiseless_low_rank}. For the under regularized case (Figure \ref{fig:noiseless_low_rank}(a)), when fixing $p,d$ and increasing $n$, the error decreases until $n=pd$, and then becomes a constant. When fixing $n,d$ and increasing $p$, the error decreases as $p^{2\gamma-1}$ and becomes a constant after $p=\frac{n}{d}$. Thus large width and lower regularization is always preferred. 

\subsection{Derivation of the excess test error for pruning}
\label{app:pruning}
Now we consider pruning, i.e., first train a full width student and only keep the leading $p\ll d$ eigenvalues. For technical simplicity, we cut the eigenvalues using $\mu\to\text{ReLU}(\mu-\mu_0)$ for some smallest $\mu_0\geq0$ such that at most $p$ eigenvalues are non-zero. In this way the excess risk is given by
\begin{equation}
Q_\star+(\delta\partial_\delta+b\partial_b-1)J(\delta,b),
\end{equation}
for some suitably chosen $b\geq\lambda\epsilon$. The phases in the following thus refer to the same phases as in \cite{defilippis2026scaling}.

For $n\ll d$ the excess test error is always $Q_\star$ and pruning has no effect. For $d\ll n\ll d^2$, the excess test error is given by the following
\begin{itemize}
\item When $\lambda\gg n/d^{3/2}$ the model is in phase I (all eigenvalues are zero). Thus pruning has no effect.
\item When $\sqrt{n/d^2}\ll\lambda\ll n/d^{3/2}$, the model is in phase II, and there are $(4n/\lambda d^{3/2})^{1/\gamma}$ spikes. Therefore, when $p\gg (4n/\lambda d^{3/2})^{1/\gamma}$ nothing is pruned, and when $p\ll (4n/\lambda d^{3/2})^{1/\gamma}$, the error scales as $p^{1-2\gamma}$.
\item When $\lambda\ll \sqrt{n/d^2}$, the model is in phases IV or V. There are $(n/d)^{1/2\gamma}$ spikes, a bulk, and other eigenvalues are zero. When $p/d\gg\int_{\tl\epsilon}^{2\delta}\mu_{sc}(x/\delta)/\delta dx=\Theta((n/d^2)^{3/5})$, nothing is pruned. When $p\ll (n/d)^{1/2\gamma}$ the error scales as $p^{1-2\gamma}$. When $(n/d)^{1/2\gamma}\ll p\ll \Theta(d(n/d^2)^{3/5})$, the overfitting error depends on
\begin{equation}
J_1(\delta,b):=\int_{b}^{2\delta}\mu_{sc}(x/\delta)/\delta(x-b)^2 dx=\delta^2\frac{16t ^{7/2}}{105\pi},
\end{equation}
where $t :=2-\frac{b}{\delta}$ satisfies $\frac{p}{d}=\int_{b}^{2\delta}\mu_{sc}(x/\delta)/\delta dx=\frac{2}{3\pi}t ^{3/2}$. Thus the total error scales as
\begin{equation}
\caE=\frac{2^{5/3} \cdot 3^{4/3} \cdot \pi^{4/3}}{35} \left( \frac{p}{d} \right)^{7/3}\frac{\Delta d^2}{4n}+\frac{24\gamma^3}{4\gamma^3+4\gamma^2-\gamma-1}\left(\frac{d\Delta}{4n}\right)^{1-\frac{1}{2\gamma}},
\end{equation}
where we use the solution $\delta^2=\frac{\Delta d^2}{4n}$ from \cite{defilippis2026scaling}. Consequently, the error is optimal (i.e. $(d/n)^{1-\frac{1}{2\gamma}}$) for $(n/d)^{1/2\gamma}\ll p\ll d^{10/7}(n/d)^{\frac{3}{14\gamma}}$. Note that we always have $(n/d)^{1/2\gamma}\ll d^{10/7}(n/d)^{\frac{3}{14\gamma}}$ for $n\ll d^{1+5\gamma}$.
\end{itemize}

For $n\gg d^2$, the excess test error is given by the following
\begin{itemize}
\item When $\lambda\gg\max(\sqrt{n/d^2},n/d^{3/2+\gamma})$, the model is in phase II. Thus when $p\gg (4n/\lambda d^{3/2})^{1/\gamma}$ nothing is pruned, and when $p\ll (4n/\lambda d^{3/2})^{1/\gamma}$, the error scales as $p^{1-2\gamma}$.
\item When $\lambda\ll n/d^{3/2+\gamma}$ and $n\gg d^{2\gamma}$, the model is in phases III or VIb. There are $d$ spikes and thus the error scales as $p^{1-2\gamma}$.
\item When $\lambda\ll\sqrt{n/d^2}$ and $d\ll n\ll d^{2\gamma}$, the model is in VIb. There are $(n/d)^{1/2\gamma}$ spikes and a bulk. When $p\ll (n/d)^{1/2\gamma}$ the error scales as $p^{1-2\gamma}$. Otherwise the error scales in the same way
\begin{equation}
\caE=\frac{2^{5/3} \cdot 3^{4/3} \cdot \pi^{4/3}}{35} \left( \frac{p}{d} \right)^{7/3}\frac{\Delta d^2}{4n}+\frac{24\gamma^3}{4\gamma^3+4\gamma^2-\gamma-1}\left(\frac{d\Delta}{4n}\right)^{1-\frac{1}{2\gamma}},
\end{equation}
as in the previous case.
\end{itemize}

\subsection{Generalization to the attention model}
\label{app:attention}

As shown by \cite{boncoraglio2026singleheadattentionhighdimensions}, the scaling phase diagram obtained for full-rank quadratic networks in \cite{defilippis2026scaling} remains very similar for the attention index model defined as
\begin{equation}
\hf_{\rm sq}(\bx_{\rm in};W) = A_W(\bx_{\rm in})\bx_{\rm in},\ \text{or}\ \hf_{\rm lb}(\bx_{\rm in};W) = A_W(\bx_{\rm in})
\end{equation}
for the seq2seq and seq2lab tasks, where $\bx_{\rm in}\in \bbR^{T \times d}$ is a sequence of $T$ tokens
\begin{equation}
A_W(\bx_{\rm in}) = \sigma\!\left(\! \frac{
        \bx_{\rm in} WW^T \bx_{\rm in}^T 
        - 
        \mathbb{E}_{\rm tr}[\bx WW^T \bx^T] 
        }{d\sqrt{p}} \right) \, \in\mathbb{R}^{T\times T}
    \label{architecture}
\end{equation}
with weights $W\in\bbR^{d \times p}$.
$\sigma:\mathbb{R}^{T\times T}\to \mathbb{R}^{T\times T}$ is the activation (e.g. softmax) and $A_W(\bx)$ is the attention matrix with tied key and query matrices. $\mathbb{E}_{\rm tr}$ refers to the empirical average over the training set.

We consider a target function that lies within the expressivity class of the architecture in Eq.~\eqref{architecture}, and restrict the class of (possibly noisy) target functions to the ones of the form (respectively for the seq2seq and seq2lab tasks)
\begin{equation}\label{eq:data1}
    \bx_{\rm out}^\mu = \sigma_*\left( R(\bx_{\rm in}) \right) \bx_{\rm in} \mathand
    y^\mu = \sigma_*\left( R(\bx_{\rm in}) \right) \,,
\end{equation}
where $R(\bx_{\rm in}) \in \bbR^{T \times T}$ is a centered
pre-activation matrix, and $\sigma_*$ is a possibly different activation function. We choose:
\begin{equation}\label{eq:data2}
     R(\bx_{\rm in})_{ab} = \frac{\bx_{\rm in, a}^T S_0 \bx_{\rm in, b} - \delta_{ab} \Tr(S_0)}{\sqrt{d}} + \sqrt{\frac{\Delta}{2 - \delta_{ab}}} \xi^\mu \, ,
\end{equation}
where $S_0$ is the target function weight matrix with eigenvalues $\{\sqrt{d}i^{-\gamma}\}_{i=1}^d$. 

We learn $W$ by {\textit{empirical risk minimization}} of the square loss with $\ell_2$ regularization (or equivalently, weight decay) that is commonly used in practice in large language models, 
i.e. $\hat {W} = \arg\min_W [\caL^{\rm data} (W) + \lambda \| W \|_F^2]$ where respectively for the two tasks
\begin{equation}
\begin{cases} 
    \caL^{\rm data}_{\rm sq}(W) \!:=\! \frac{1}{d} \sum_{\mu=1}^n ||\bx_{\rm out}^\mu - \hf_{\rm sq}(\bx_{\rm in}^\mu;W)||_F^2 \, ,
    \\
    \caL^{\rm data}_{\rm lb}(W) \!:=\! \sum_{\mu=1}^n ||y^\mu - \hf_{\rm lb}(\bx_{\rm in}^\mu;W)||_F^2 \, .
\end{cases}
\label{eq:erm}
\end{equation}
We measure the performance of the learned network through the test errors 
\begin{equation}\label{eq:test}
\begin{aligned}
&e_{\rm test} = \frac{1}{d} \mathbb{E}_{\bx_{\rm in}, \bx_{\rm out}} || \bx_{\rm out} - \hf(\bx_{\rm in};\hat{W}) ||_F^2,\\
&e_{\rm test} = \mathbb{E}_{\bx_{\rm in}, y} || y - \hf(\bx_{\rm in};\hat{W}) ||_F^2 \, ,
\end{aligned}
\end{equation}
where $\mathbb{E}$ stands for an average over an appropriate test set.

Then we conjecture that the following excess test error satisfies the same phase diagram as in the main text
\begin{equation}
\caE:=e_{\rm test}-e_{\rm base},
\end{equation}
where
\begin{equation}
e_{\rm base}:=\inf_{m} \mathcal{I}(m_0,m_0/Q_\star)
\end{equation}
for the interpolation regime ($\tl\ll\sqrt{n/d^2}$ and $n\ll dp$ such that the training loss is vanishing and the minimum is flat) and
\begin{equation}
e_{\rm base}:=\inf_{m} \mathcal{I}(m,m/Q_\star)
\end{equation}
otherwise. Here we define
\begin{equation}
\mathcal{I}(m,q)=\mathbb{E}_{z_0,z,\xi}||\tilde{\sigma}_*(z_0+\sqrt{\Delta/2}\zeta)-\tilde{\sigma}(z)||^2,
\end{equation}
where $(z_0,z)\sim\mathcal{N}\left(0,\left(\begin{array}{cc}
    Q_0 &m  \\
     m&q 
\end{array}\right)\right)$, $\zeta\sim\mathcal{N}(0,1)$. $\tilde{\sigma}_*$ is the effective teacher activation and $\tilde{\sigma}$ is the effetive student activation, defined as $\tilde{\sigma}(A) := \sigma(\{\sqrt{1 + \delta_{ab}} \, A_{ab}\}_{ab})$ for $A \in \bbR^{T \times T}$ and similarly $\tilde{\sigma}_*(A) := \sigma_*(\{\sqrt{1 + \delta_{ab}} \, A_{ab}\}_{ab})$.

Finally $m_0:=\eta Q_\star$ is defined as the solution of the following equation
\begin{equation}
\eta=-\frac{1}{2}\left[\frac{\partial \mathcal{K}(m,q)}{\partial q}|_{\eta Q_\star,\eta^2 Q_\star}\right]^{-1}\frac{\partial \mathcal{K}(m,q)}{\partial m}|_{\eta Q_\star,\eta^2 Q_\star},
\end{equation}
where 
\begin{equation}
\mathcal{K}(m,q):=\mathbb{E}_{z_0,z,\zeta}\inf_{h\in\arg\inf||\tilde{\sigma}_*(z_0+\sqrt{\Delta/2}\zeta)-\tilde{\sigma}(\cdot)||^2}||h-z||^2
\end{equation}
with $\zeta\sim\mathcal{N}(0,1)$ independent of $(z_0,z)$.

\section{Details of numerical implementations}
\label{app:numerics}

Here we describe our numerical implementation. The code can be found in the repository \url{https://github.com/SPOC-group/WidthQuadraticNetworks}. 

Appendix \ref{app:numerics_theory} shows how we solved the system \eqref{eq:SE_ERM} numerically to obtain the theoretical predictions for our figures, and are always plotted with continuous lines. 

Appendix \ref{app:numerics_exp} describes how we conducted the experiments, which we draw using dots with error bars indicating the standard deviation among at least two realizations for each value of the parameters. In all the plots that we show, the error bars are too small to be distinguished to the markers, we thus attribute the tiny deviations from the analytical theory either to failures of the training algorithm or to finite size effects that are expected to disappear by increasing $d$. 

We include a working code to reproduce our results.
The theory can be reproduced on a modern laptop. We ran our implementation on a MacBook Pro with Apple M4 Pro and 24/48GB of RAM. Some of the experiments require large memory, and they were conducted on a cluster with up to 1TB of RAM (56k CPU hours).

\subsection{Theoretical lines} \label{app:numerics_theory}
To plot theoretical predictions, we solve \eqref{eq:SE_ERM} in the form \eqref{eq:amp-se}, i.e.
\begin{equation}
\begin{aligned}
\begin{cases}
\displaystyle
\widehat{\Sigma}
= \frac{2n}{d^2}\,\frac{1}{\Sigma + \frac14},
\\[0.75em]
\displaystyle
\widehat{m}
= \frac{2n}{d^2}\,\frac{1}{\Sigma + \frac14},
\\[0.75em]
\displaystyle
\widehat{q}
= \frac{2n}{d^2}\,
\frac{Q_\star - 2m + q + \frac{\Delta}{2}}
{\left(\Sigma + \frac14\right)^2},
\end{cases}
\qquad
\begin{cases}
\displaystyle
m
= -2\,\partial_{\widehat{m}}
\Psi(\widehat{\Sigma},\widehat{q},\widehat{m}),
\\[0.75em]
\displaystyle
q
= 4\,\partial_{\widehat{\Sigma}}
\Psi(\widehat{\Sigma},\widehat{q},\widehat{m}),
\\[0.75em]
\displaystyle
\Sigma
= -4\,\partial_{\widehat{q}}
\Psi(\widehat{\Sigma},\widehat{q},\widehat{m}) .
\end{cases}
\end{aligned}
\end{equation}
with 
\begin{equation}
\begin{split}
    \Psi(\hSigma, \hq, \hm) &= 
         - \frac{\hm^2}{4\hSigma} J_p\left( \frac{\sqrt{\hq}}{\hm}, \frac{2\tl}{\hm}\right)
     \, .
\end{split}
\end{equation} 
We found that taking explicit derivatives of $\Psi$ is numerically unstable for power-law targets, and for this reason we resorted to the solution of the equivalent set of equations

\begin{equation}\label{eq:app:final-SE}
    \begin{split}
        \begin{cases}
            m&
            = \frac{1}{d}{\mathbb E}_{\bhS,\bS_\star} \left[\Tr(\bhS^\top\bS_\star )\right]\\
            q&= \frac{1}{d}{\mathbb E}_{\bhS,\bS_\star} \left[
             \Tr(\bhS^\top\bhS )\right]\\
            \Sigma &= \frac{2}{d}\bbE_\bhS\left[\sum_{i=1}^{p}\frac{\Theta(\nu_i)}{\Hat{\Sigma}^t}+\sum_{i<j}\frac{\Tilde{\nu}_i-\Tilde{\nu}_j}{\nu_i-\nu_j}\right]
        \end{cases}
    \end{split}
\end{equation}
where $\bhS= (\bS_\star + \delta \bZ - \tl \epsilon \bI_d)^+_{(p)} $,  $\nu_i$ is the $i^\mathrm{th}$ eigenvalue of the  matrix $(\bS_\star + \delta \bZ - \tl \epsilon \bI_d)$ and $\Tilde{\nu}_i$ is the $i^\mathrm{th}$ eigenvalue of $\bhS$.
These equations can be recast into the form \eqref{eq:SE_ERM} by defining $\delta=\sqrt{\hq}/\hm$ and $\epsilon=2/\hm$.
The alternative expression for $m,q,\Sigma$ can be found in \cite[Appendix A.4.3]{erba2025nuclear}, and is discussed in Appendix \ref{app:derivation_SE}. The expression for $\Sigma$ is obtained computing the divergence \cite[Eq 54]{erba2025nuclear} explicitly in terms of eigenvalues.

In practice, we compute ${\mathbb E}_{\bhS,\bS_\star}$ by taking $\bS_\star$ deterministic with power-law diagonal (as defined in the main text), by sampling $n_{\rm empirical}$ GOE noise matrices $\bZ\sim GOE(d)$, computing for each of the GOE matrices $(\bhS, \nu_i, \Tilde{\nu}_i)$, and taking the empirical means of the resulting $(m, q, \Sigma)$ variables. In practice, we used $n_{\rm empirical} = 16$ in our plots.
We then iterate by fixed point scheme \eqref{eq:app:final-SE} until convergence. Crucially, we keep the same GOE matrices for all the iterations of the fixed point scheme in order to avoid unnecessarily long solution time.

\subsection{Experiments} \label{app:numerics_exp}
To run experiments on this architecture, we use the solver LBFGS with the PyTorch implementation \cite{paszke2019pytorch} to minimize the objective \eqref{eq:def:erm}. The specific hyperparameters used can be found in the code provided. As we remarked in the main text, the problem is in non-convex, so we are not guaranteed to converge to the global minimum. Despite this, we observe a remarkably good agreement with our theory.
\subsection{Additional figures}
Here some additional figures that show the match between our theory and experiments. In particular, in the regime with decay $\caE=\Theta(d^{-\rho(1-2\gamma)})$, where we expect to have enough samples to perfectly recover the dominant teacher eigenspaces, we compare our results to the lower bound provided by the Eckart-Young theorem \citep{Eckart_Young_1936} for our target structure (eigenvalues $\{\sqrt{d/\zeta(2\gamma)}i^{-2\gamma}\}_{i=1}^d$)
\begin{equation}\label{eq:EY}
    \caE(p)\geq\caE_{EY}(p)=\frac{1}{\zeta(2\gamma)} \sum_{i=p+1}^d i^{-2\gamma}
\end{equation}

\begin{figure}
    \centering
    \includegraphics[width=0.6\linewidth]{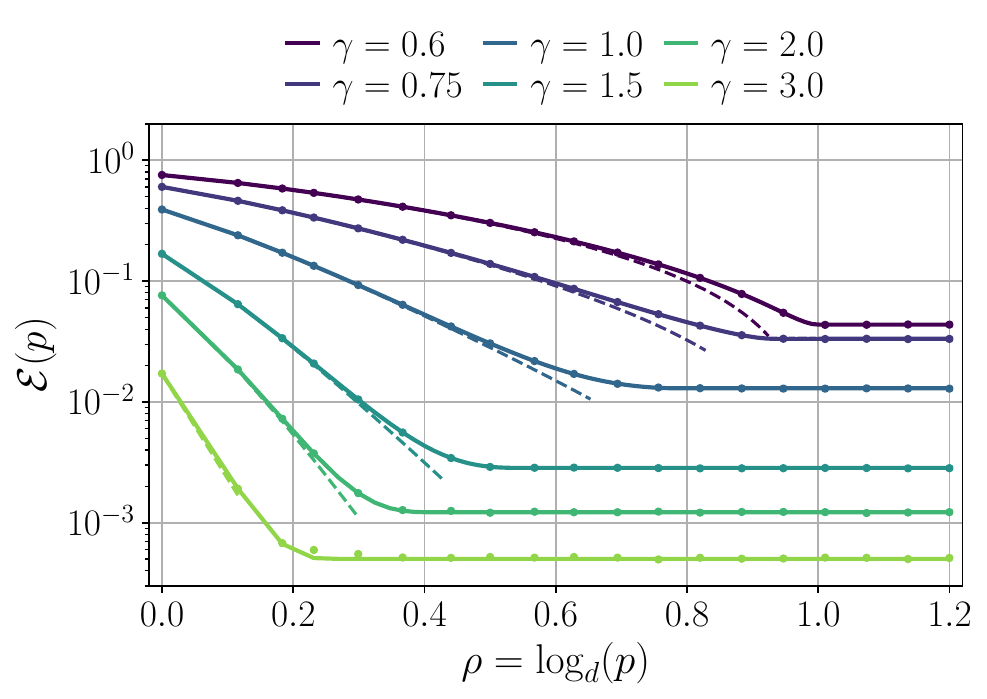}
    \caption{Test error scaling in the over-regularized regime for multiple data exponents $\gamma$, with $d=400$, $\alpha=2.5$, $\sqrt{\Delta}=0.5$ and $\ell=0.45$. Full lines are non-asymptotic state evolution, dots are experiments (LBFGS) and dashed lines are the low rank estimation lower bound \eqref{eq:EY}.}
    \label{fig:fig3_verif}
\end{figure}

\begin{figure}
    \centering
    \includegraphics[width=\linewidth]{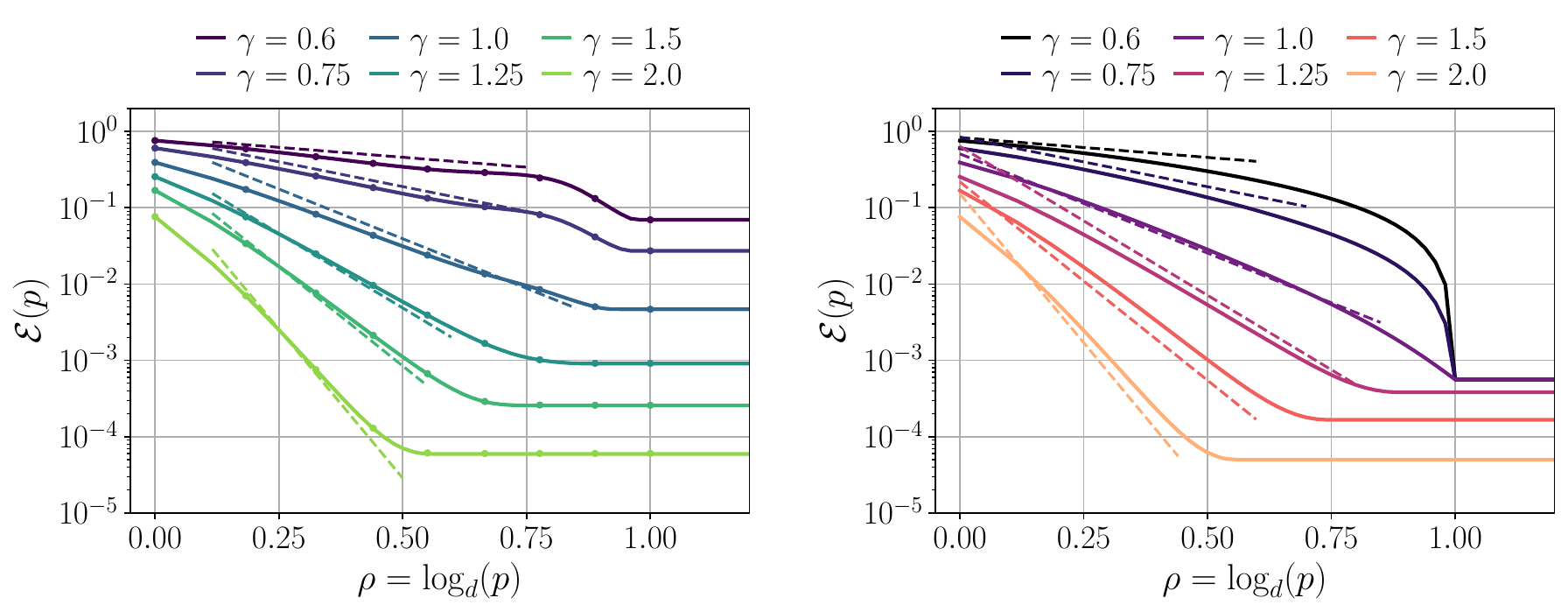}
    \caption{(Left) Test error scaling behavior in the under-regularized noiseless regime for different values of $\gamma$ as a function of the width $p$, for fixed $d=400$, $\alpha=1.9$ and $\ell= -0.5$. 
    (Right) Test error scaling behavior in the over-regularized noiseless regime for different values of $\gamma$ as a function of the width $p$, for fixed $d=400$, $\alpha=2.5$ and $\ell= 0.1$. }
        \label{fig:fignoiseless}
\end{figure}

\begin{figure}
    \centering
    \includegraphics[width=1.0\linewidth]{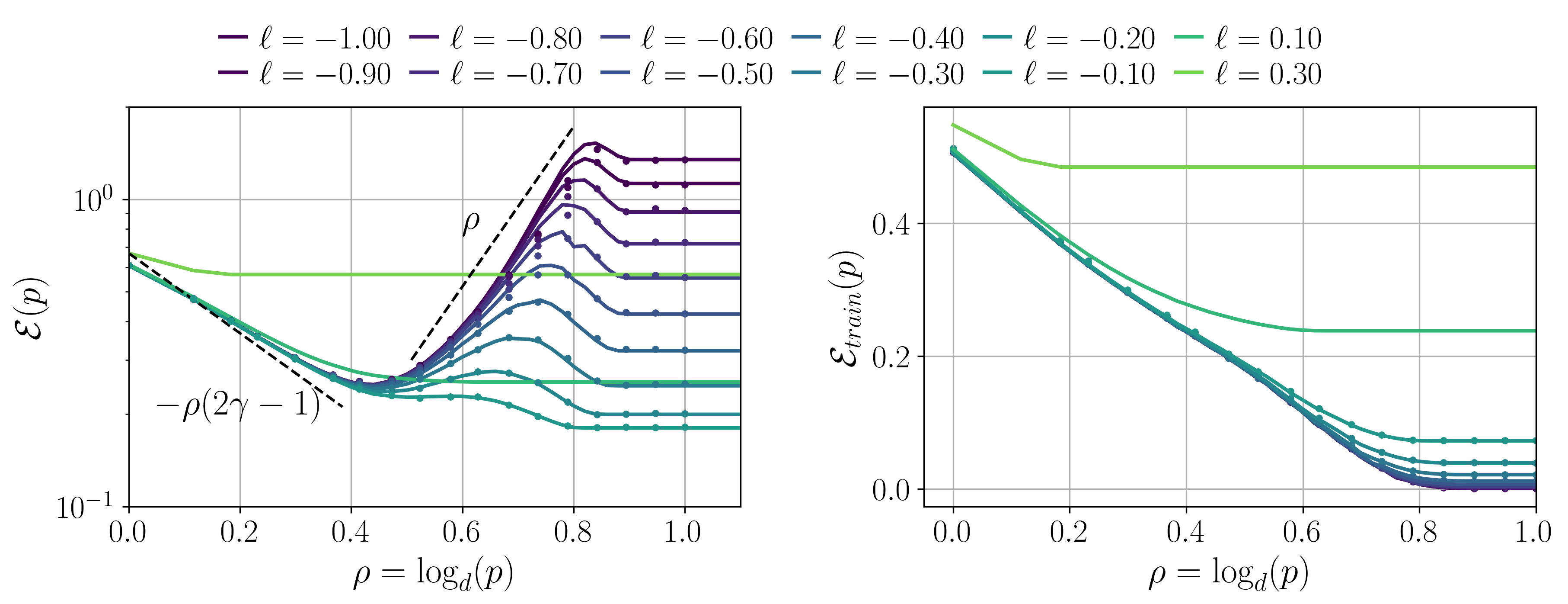}
    \caption{Test error (left) and training error (right) as functions of the network width for different regularization strengths $\ell$. Parameters are $d=400$, $\alpha=1.8$, $\sqrt{\Delta}=0.5$, and $\gamma=0.75$. Solid lines correspond to the non-asymptotic state evolution predictions, while dots denote LBFGS experiments. The dashed lines show the corresponding scalings with $\rho$ from Result \ref{res:under-reg}. As the regularization is decreased, a double-descent phenomenon emerges. Increasing the width initially reduces both the training and test errors. Near the interpolation regime, the training error reaches a plateau while the test error exhibits a pronounced peak. Further increasing the width maintains a low training error and subsequently improves generalization performance.}
    \label{fig:figDD}
\end{figure}

\section{Declaration of LLM usage} \label{app:LLMusage}
We used LLMs in the following stages of the preparation of this work: editing (e.g., grammar, spelling, word choice), drafting sections of the paper, facilitating or running experiments, visualizing results for submission. Every LLM output was double checked manually by the authors to ensure it performs as intended, in particular with regards to the numerical experiments.

\end{document}